%% file: arxiv.tex
\documentclass[lettersize,journal,compsoc]{IEEEtran}
\usepackage{amsmath,amsfonts}
\usepackage{algorithmic}

\usepackage{graphicx}
\usepackage{adjustbox}

\usepackage{hyperref}
\usepackage{cite}
\usepackage[caption=false,font=normalsize,labelfont=sf,textfont=sf]{subfig}
\usepackage{url}

\usepackage{lmodern}
\usepackage{amsfonts}

\usepackage{textcomp}
\usepackage{stfloats}
\usepackage{subcaption}
\usepackage[table]{xcolor}
\usepackage[most]{tcolorbox}
\usepackage{caption}

\usepackage{verbatim}
\usepackage{graphicx}
\usepackage{booktabs}
\usepackage{tabularx}
\usepackage{makecell}
\usepackage{array}
\usepackage{fontawesome}
\usepackage{multirow}
\usepackage{xcolor}

\usepackage{graphicx}

\def\BibTeX{{\rm B\kern-.05em{\sc i\kern-.025em b}\kern-.08em
    T\kern-.1667em\lower.7ex\hbox{E}\kern-.125emX}}
\usepackage{balance}

\begin{document}

\title{\includegraphics[scale=0.04]{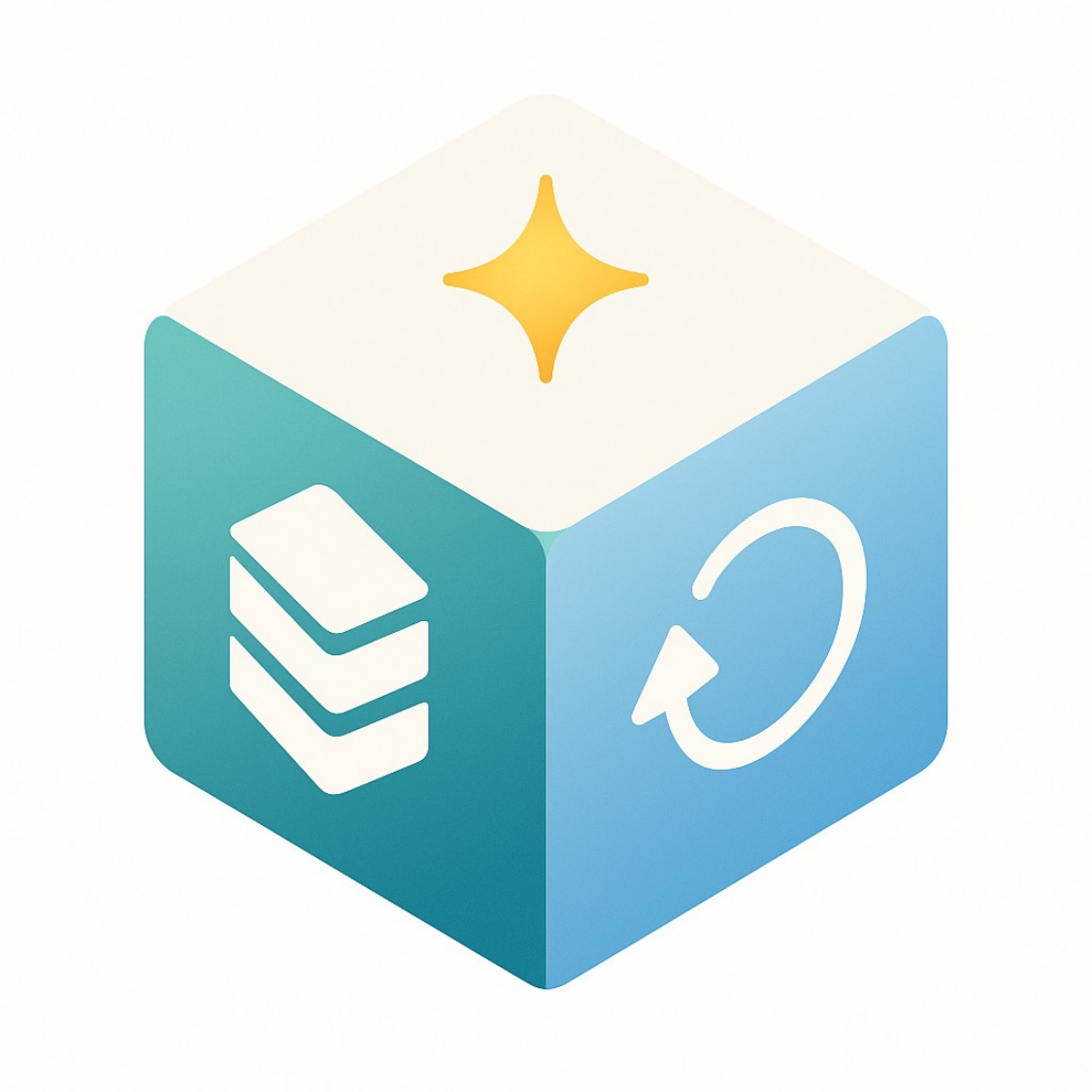}R\textsuperscript{3}eVision: A Survey on Robust Rendering, Restoration, and Enhancement \\for 3D Low-Level Vision}

\author{Weeyoung~Kwon,
        Jeahun~Sung,
        Minkyu~Jeon,
        Chanho~Eom,
        and Jihyong Oh\textsuperscript{\dag}
\IEEEcompsocitemizethanks{
\IEEEcompsocthanksitem
Weeyoung~Kwon and Chanho Eom are with the Department of Metaverse Convergence, GSAIM, Chung-Ang University, Seoul, South Korea (e-mail: weeyoungkwon@cau.ac.kr; cheom@cau.ac.kr).
\IEEEcompsocthanksitem
Minkyu~Jeon is with the Department of Computer Science, Princeton University, Princeton, NJ, USA (e-mail: mj7341@princeton.edu).
\IEEEcompsocthanksitem
Jeahun~Sung and Jihyong Oh are with the Department of Imaging Science, GSAIM, Chung-Ang University, Seoul, South Korea (e-mail: jhseong@cau.ac.kr; jihyongoh@cau.ac.kr).
}
\thanks{\textsuperscript{\dag} denotes corresponding author.}%
}

\input{sections/abstract}

\maketitle
\IEEEdisplaynontitleabstractindextext

\input{sections/introduction}
\input{sections/Problem_Definition_and_Challenges}
\input{sections/Preliminaries}
\input{sections/Low-Level_Vision_for_Robust_3D_Rendering}
\input{sections/Experimental_Setup}
\input{sections/Future_Directions}
\input{sections/Conclusion}

\bibliographystyle{IEEEtran}

\bibliography{ref}

\vspace*{-10em}
\newcounter{dummy}

\refstepcounter{dummy}\label{ref188}

\refstepcounter{dummy}\label{ref189}

\refstepcounter{dummy}\label{ref190}
\nocite{b210}
\nocite{b211}
\nocite{b212}
\end{document}

%% file: sections/abstract.tex
\IEEEtitleabstractindextext{%
\begin{abstract}
Neural rendering methods such as Neural Radiance Fields (NeRF) and 3D Gaussian Splatting (3DGS) have achieved significant progress in photorealistic 3D scene reconstruction and novel view synthesis. However, most existing models assume clean and high-resolution (HR) multi-view inputs, which limits their robustness under real-world degradations such as noise, blur, low-resolution (LR), and weather-induced artifacts. To address these limitations, the emerging field of 3D Low-Level Vision (3D LLV) extends classical 2D Low-Level Vision tasks including super-resolution (SR), deblurring, weather degradation removal, restoration, and enhancement into the 3D spatial domain. This survey, referred to as R\textsuperscript{3}eVision, provides a comprehensive overview of robust rendering, restoration, and enhancement for 3D LLV by formalizing the degradation-aware rendering problem and identifying key challenges related to spatio-temporal consistency and ill-posed optimization. Recent methods that integrate LLV into neural rendering frameworks are categorized to illustrate how they enable high-fidelity 3D reconstruction under adverse conditions. Application domains such as autonomous driving, AR/VR, and robotics are also discussed, where reliable 3D perception from degraded inputs is critical. By reviewing representative methods, datasets, and evaluation protocols, this work positions 3D LLV as a fundamental direction for robust 3D content generation and scene-level reconstruction in real-world environments. We maintain an up-to-date project page: \url{https://github.com/CMLab-Korea/Awesome-3D-Low-Level-Vision}.
\end{abstract}
\begin{IEEEkeywords}
3D Computer Vision, Low-Level Vision, Neural Rendering, Restoration, Enhancement, NeRF, 3DGS
\end{IEEEkeywords}
}

%% file: sections/introduction.tex
\ifCLASSOPTIONcompsoc
  {\IEEEraisesectionheading{\section{Introduction}\label{sec:intro}}}
\else
  {\section{Introduction}\label{sec:intro}}
\fi
\IEEEPARstart{V}ision is one of the most important human senses, enabling people to perceive and interact with the three-dimensional (3D) world in which they live. Understanding and representing 3D space is essential for applications such as the metaverse, augmented reality (AR), virtual reality (VR), robotics, and immersive media. Various approaches to 3D representation have received significant attention, as reflected in the rapid development of the field and the growing body of related work \cite{b3}.

The emergence of recent rendering frameworks such as Neural Radiance Fields (NeRF) \cite{b1} and 3D Gaussian Splatting (3DGS) \cite{b2} has led to significant progress in photorealistic 3D scene reconstruction and novel view synthesis \hyperref[ref189]{[189]}.

\begin{figure}[t]
    \centering
    \includegraphics[width=0.85\linewidth]{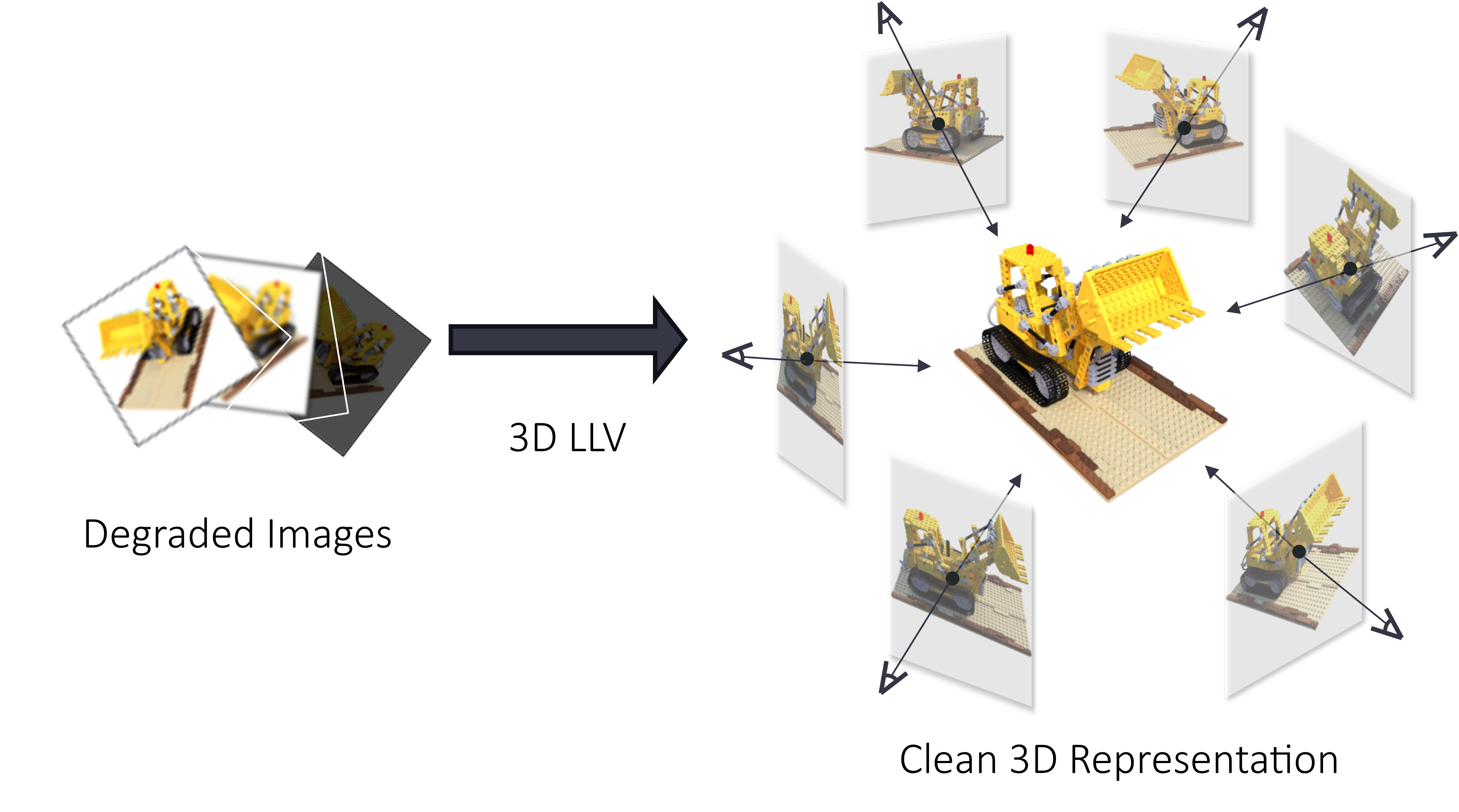}
    \caption{Overview of 3D LLV tasks, including Super-Resolution (SR), Deblurring, Weather Degradation Removal, Enhancement, and Restoration. These tasks aim to recover high-quality (HQ) 3D representations from degraded or sparse observations while maintaining view-consistent fidelity under real-world degradations.}
    \label{fig:3DLLV}
\end{figure}

Despite recent advances in neural rendering, most existing models assume clean and high-Resolution (HR) multi-view observations~\cite{b78, b85, b71, b76}, which makes them susceptible to real-world degradations such as LR, noise, weather artifacts, and blur~\cite{b99, b98}. As shown in Fig. \hyperref[fig:3DLLV]{1}, these degradations frequently lead to structural distortions in 3D reconstruction and inconsistencies across viewpoints.
\begin{figure*}[t]
    \centering
    \includegraphics[width=0.8\textwidth]{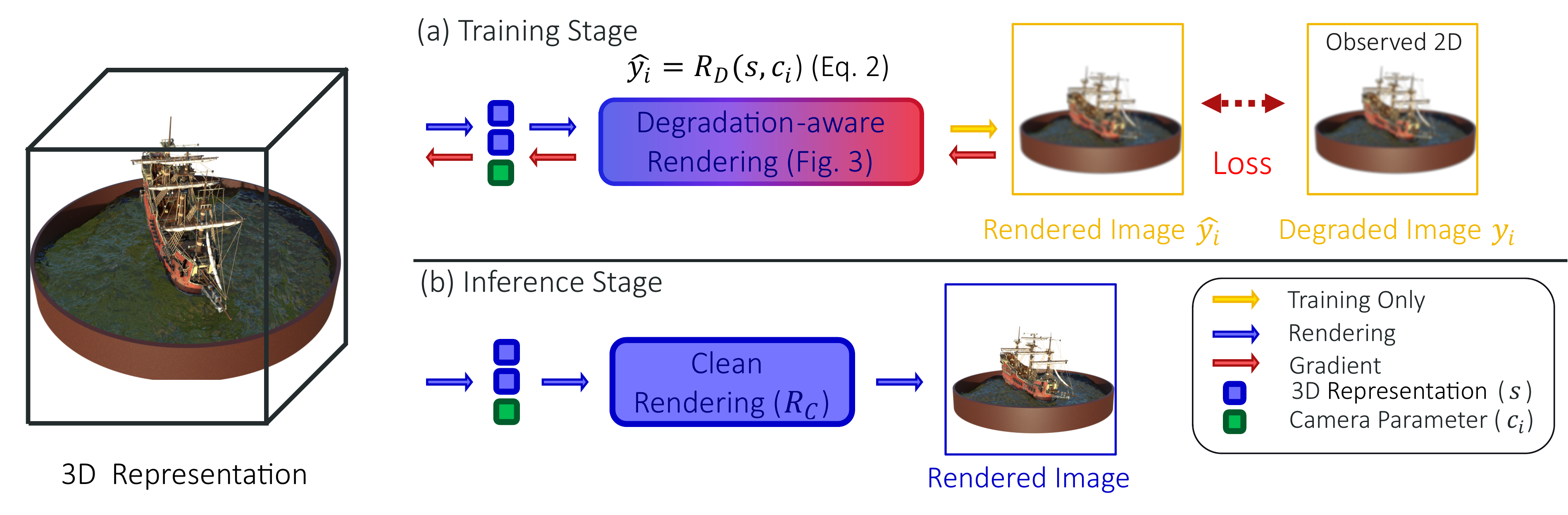}
    \caption{Overview of the degradation-aware rendering pipeline for 3D Low-Level Vision (3D LLV).
    During the training stage (a), the model generates degraded renderings \( \hat{y}_i = R_D(s, c_i) \) using a degradation-aware rendering function \( R_D \), conditioned on the 3D representation \( s \) and camera parameters \( c_i \), as defined in Eq. \hyperref[eq:degradation-aware-rendering]{2}.
    The degradation modeling follows either sequence-based rendering, where clean images are rendered and subsequently degraded by view-specific operators, or composition-based rendering, where clean and degraded renderings are fused through a differentiable function, as shown in Fig.~\ref{fig:degradation-modeling}.The loss is computed between the rendering \( \hat{y}_i \) and the observed degraded input \( y_i \).
    During inference (b), the learned clean 3D representation \( s \) is used to generate high-quality images via the clean renderer \( R_c \).
  }

    \label{fig:degradation-aware-rendering}
\end{figure*}
To address these limitations, 3D Low-Level Vision (3D LLV) has recently emerged as a prominent research direction. 3D LLV extends classical LLV tasks such as Super-Resolution (SR), Deblurring, Weather Degradation Removal, Enhancement, and Restoration into the 3D spatial domain. The goal is to reconstruct high-fidelity 3D content from degraded 2D observations while ensuring 3D spatio-temporal consistency throughout the reconstruction process~\cite{b97, b13}.

In 2D LLV, degradations are commonly modeled using the following equation:
\begin{equation}
\mathbf{y} = \mathbf{H} \mathbf{x} + \mathbf{n},
\end{equation}
where \( \mathbf{y} \) denotes the observed degraded 2D image, \( \mathbf{x} \) is the clean 2D image, \( \mathbf{H} \) is a 2D degradation operator (e.g., blur, downsampling), and \( \mathbf{n} \) represents additive noise. The objective is to recover \( \mathbf{x} \) from a single degraded input \( \mathbf{y} \), focusing on pixel-wise enhancement or restoration.

In contrast, 3D LLV aims to reconstruct a geometrically and temporally consistent 3D scene representation \( s \) from a set of \( N \) observed multi-view degraded images \( \{ y_i \}_{i=1}^{N} \), each associated with camera parameters \( \{ c_i \}_{i=1}^{N} \). The rendering process in this context must account for both the underlying 3D geometry and view-specific degradations. To model this, we define the rendered degraded view \( \hat{y}_i \) as:
\begin{equation}
\hat{y}_i = R_D(s, c_i), \quad D \in \{D_{\text{seq}}, D_{\text{comp}}\},
\label{eq:degradation-aware-rendering}
\end{equation}
where \( \hat{y}_i \) denotes the rendered degraded image, \( R_D \) is the degradation-aware rendering function conditioned on both the camera pose \( c_i \) and the chosen degradation strategy \( D \in \{D_{\text{seq}}, D_{\text{comp}}\} \). An optional additive noise term \( n_i \) may also be implicitly included to account for sensor-level perturbations. Eq.\ref{eq:degradation-aware-rendering} is illustrated in Fig.~\hyperref[fig:degradation-aware-rendering]{2(a)}, which depicts the training stage where the degradation-aware rendering function \( R_D \) generates \( \hat{y}_i \) to be supervised against the observed degraded image \( y_i \). In contrast, Fig.~\hyperref[fig:degradation-aware-rendering]{2(b)} shows the inference stage, where a clean renderer \( R_c \) is used to generate high-quality outputs from the learned 3D representation \( s \) without applying degradation.

\begin{figure}[t]
    \centering
    \includegraphics[width=0.7\linewidth]{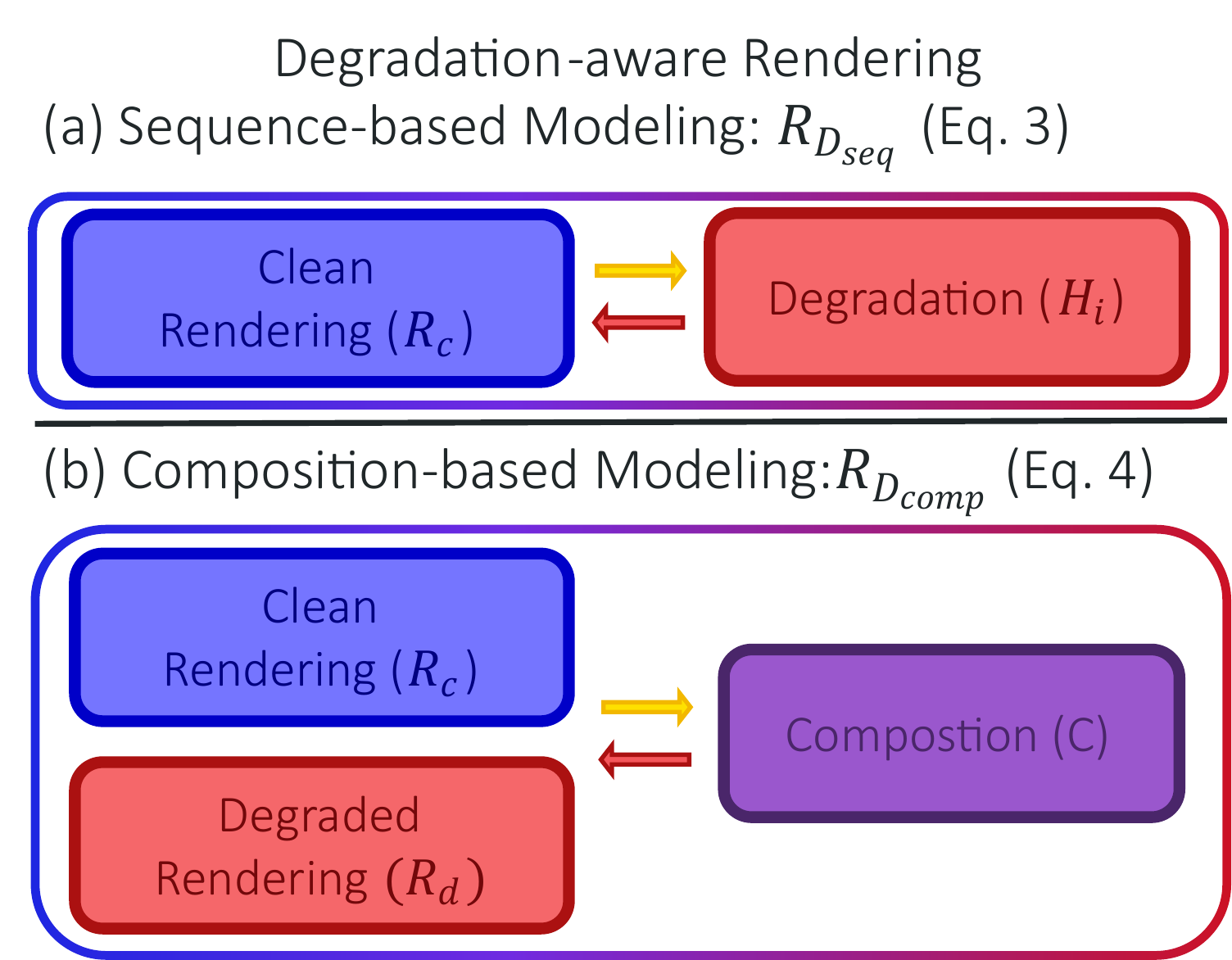}
    \caption{Comparison of two modeling approaches for Degradation-aware Rendering.
    (a) Sequence-based modeling (denoted as $R_{D_\mathrm{seq}}$): Applies a degradation function after rendering to simulate degraded observations.
    (b) Composition-based modeling (denoted as $R_{D_\mathrm{comp}}$): Combines clean and degraded renderings via a physically-based composition process to generate the final output.}

    \label{fig:degradation-modeling}
\end{figure}

We distinguish between two types of degradation modeling in 3D LLV: sequence-based modeling $R_{D_\mathrm{seq}}$ and composition-based modeling $R_{D_\mathrm{comp}}$.
First, in sequence-based modeling, the clean image is rendered using a clean renderer and subsequently degraded through a view-specific operator that models effects such as blurring or downsampling, as shown in Fig. \hyperref[fig:degradation-modeling]{3 (a)}.
Formally, the rendered degraded view for camera \( c_i \) is given by:
\begin{equation}
R_{D_{\text{seq}}}(s, c_i) = H_i R_c(s, c_i),
\end{equation}
where \( R_c(s, c_i) \) denotes the clean rendering and \( H_i \) is a degradation operator specific to view \( i \).
Second, in composition-based modeling, clean and degraded renderings are generated independently and then fused using a differentiable composition function, as shown in Fig. \hyperref[fig:degradation-modeling]{3 (b)}.
Formally, the rendered degraded view for camera \( c_i \) is defined as:
\begin{equation}
R_{D_{\text{comp}}}(s, c_i) = \mathcal{C}(R_c(s, c_i), R_d(s, c_i)),
\end{equation}
where \( R_c(s, c_i) \) and \( R_d(s, c_i) \) denote the clean and degraded renderings, respectively, and \( \mathcal{C} \) is a differentiable composition operator that fuses the two outputs.

 Eq. \ref{eq:degradation-aware-rendering} highlights that 3D LLV is inherently more ill-posed than 2D LLV, as only degraded observations \( y_i \) are available and the goal is to reconstruct a consistent and high-quality 3D representation \( s \). The predictions \( R_D(s, c_i) \) must satisfy both geometric consistency across views and temporal coherence across frames. Compared to conventional 2D LLV, this increases the complexity of the problem by requiring scene-level reasoning and multi-view regularization.

Therefore, 3D LLV demands a shift from local pixel-based restoration to globally consistent scene-level rendering that explicitly incorporates 3D geometry and degradation priors. We identify these two degradation-aware strategies as fundamental modeling approaches for robust neural rendering in real-world conditions.

Conventional 3D reconstruction and neural rendering methods are often vulnerable to degraded inputs. In contrast, 3D LLV facilitates the integration of restoration modules into frameworks such as NeRF and 3DGS, thereby enabling high-fidelity 3D scene reconstruction under conditions of noise, blur, and LR. As a result, 3D LLV has emerged as a key research direction for achieving robust and reliable 3D perception in real-world environments~\cite{b99, b98, b100}.

In autonomous driving \cite{b140,b141}, accurate reconstruction of 3D spatial information from noisy sensor inputs under varying weather, lighting, and environmental conditions is essential for tasks such as object detection, distance estimation, and path planning, which highlights the necessity of reliable 3D recovery.

In AR/VR and metaverse environments \cite{b142,b143}, \hyperref[ref188]{[188]} realistic 3D content must be generated in real time in response to changes in user viewpoint, even under input constraints such as weather degradation or LR conditions. This requires advanced restoration pipelines capable of preserving structural integrity.

In robotics \cite{b144,b145,b146}, where real-time perception and decision-making are required, it is critical to reconstruct precise 3D representations of the surrounding environment from sparse and LR sensor data to enable reliable navigation, manipulation, and interaction. Preserving fine structural details, even under degraded inputs, is of paramount importance.

To meet these demands, there is an urgent need for 3D LLV frameworks capable of producing HQ 3D reconstructions and representations under adverse conditions, including LR, weather degradation, and blur. This paper aims to investigate the emerging domain of 3D LLV and to examine how such approaches are being deployed across a diverse range of application areas.

To this end, Section \hyperref[Problem]{2} defines the typical types of degradation observed in real-world scenarios and outlines the core challenges they pose for 3D neural rendering systems. Section \hyperref[PRELIMINARIES]{3} introduces the foundational principles of representative neural rendering methods, focusing on NeRF and 3DGS. Section \hyperref[Low-Level]{4} surveys recent advances in LLV processing, including SR, deblurring, weather degradation removal, enhancement, and restoration, and highlights how these techniques are integrated into 3D rendering pipelines. Section \hyperref[Dataset]{5} presents widely used datasets and evaluation metrics for benchmarking performance. Section \hyperref[Future]{6} discusses future research directions. Section \hyperref[Conclusion]{7} concludes the paper with a summary of key contributions.

To the best of our knowledge, this is the first comprehensive survey dedicated to 3D LLV. We systematically categorize and review core methodologies that address fundamental LLV challenges in 3D space, including SR, Deblurring, Weather Degradation Removal, Enhancement, and Restoration. By systematically categorizing representative methods, datasets, and evaluation protocols, this paper establishes a comprehensive understanding of current advancements and identifies key limitations in 3D LLV research. We position 3D LLV as a foundational direction for reconstructing and representing high-fidelity 3D structures from degraded visual inputs, and establish its significance as an integral component of future 3D content generation and scene-level reconstruction.

%% file: sections/Problem_Definition_and_Challenges.tex
\section{Problem Definition and Challenges} \label{Problem}

\begin{figure}[t]
    \centering
    \includegraphics[width=0.6\linewidth]{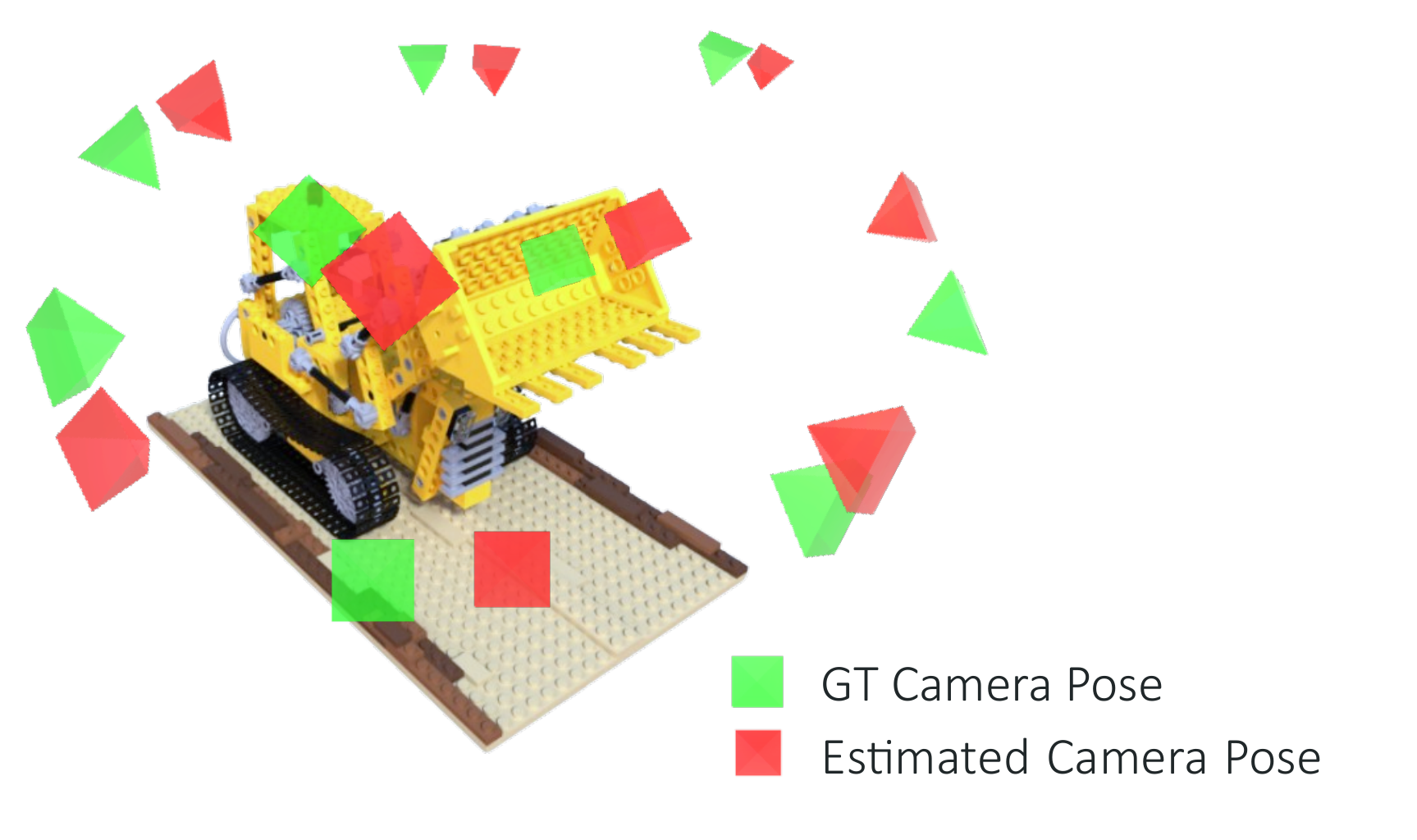}
    \caption{Comparison between GT and estimated camera poses in multi-view settings. Green frustums indicate GT camera poses, while red frustums represent inaccurate poses estimated by SfM under degraded observations. Visual degradation leads to pose estimation errors and inconsistency in 3D reconstruction.}
    \label{fig:camera}
\end{figure}

Neural rendering methods based on NeRF \cite{b1} and 3DGS \cite{b2} have demonstrated impressive performance in reconstructing photorealistic 3D scenes from HR and HQ images. However, real-world visual data are often subject to various forms of degradation, which can significantly impair reconstruction accuracy and spatiotemporal consistency~\cite{b69, b88}. In particular, Structure-from-Motion (SfM)~\cite{b16, b17, b18} and COLMAP~\cite{b14, b15}, widely used for multi-view camera pose estimation, are highly sensitive to degraded inputs. In the presence of blur, noise, LR, or illumination inconsistencies in the given images, feature detection and matching become unreliable, frequently resulting in inaccurate or failed pose estimation~\cite{b78, b171}. As shown in Fig. \hyperref[fig:camera]{4}, such degradation causes significant deviations between the estimated and GT camera poses, which in turn propagate through the rendering pipeline. Since these poses serve as critical initialization for methods like NeRF and 3DGS, any misalignment can severely degrade the final reconstruction quality.

These challenges stem from various types of visual degradation commonly encountered in real-world scenarios, and understanding how each type affects 3D neural rendering is essential for improving the robustness and reconstruction quality of methods such as NeRF and 3DGS. To this end, we categorize representative degradation types and analyze their impact on rendering pipelines.

\begin{figure*}[t]
    \centering
    \includegraphics[width=0.9\textwidth]{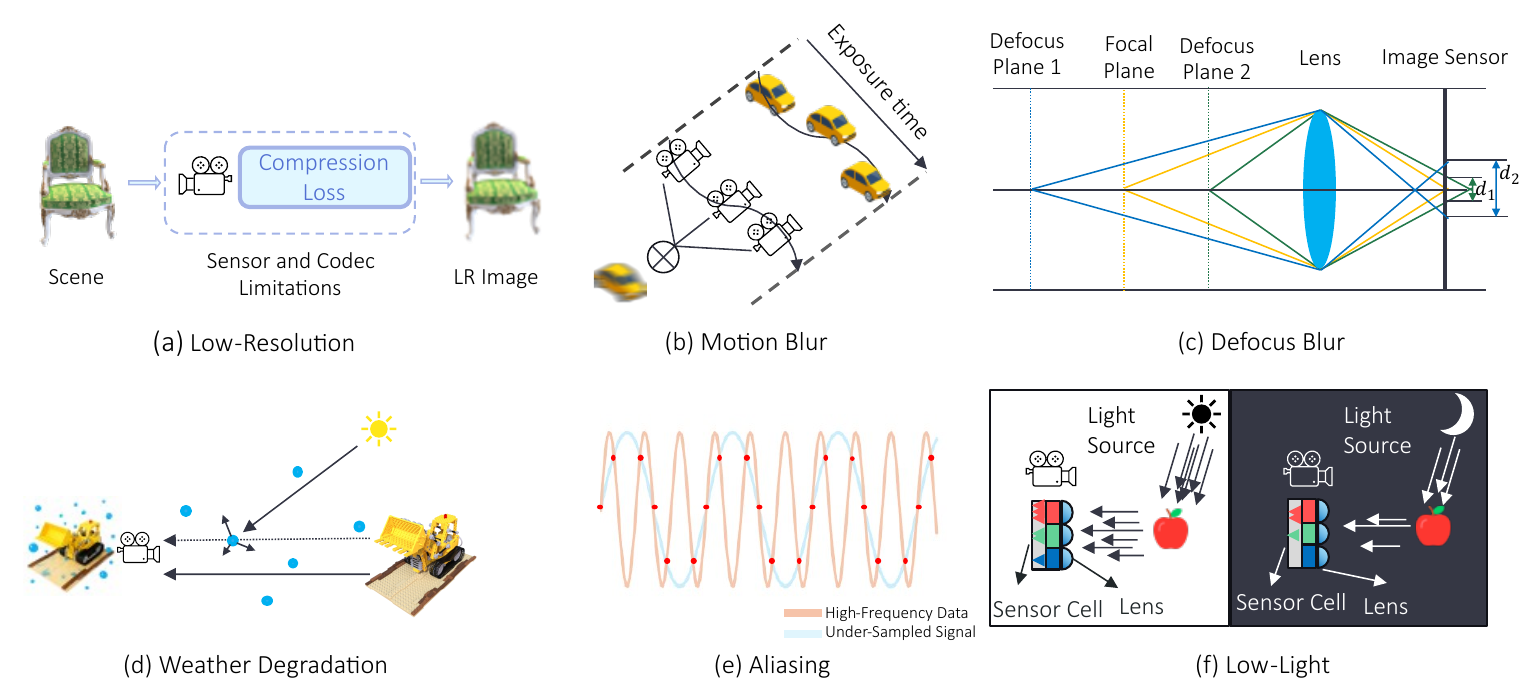}
    \caption{
    Illustration of representative visual degradation factors that cause performance degradation in neural rendering systems.
    (a) Acquisition and compression degradation, where sensor resolution limits and lossy compression suppress high-frequency components, reducing image sharpness.
    (b) Motion blur caused by temporal integration of spatial information due to object or camera movement during exposure time.
    (c) Defocus blur resulting from rays diverging on the sensor due to deviations from the focal plane. $d_1$ and $d_2$ denote the near and far points of the depth-of-field (DOF) range, respectively.
    (d) Weather degradation, where light scattering and attenuation caused by fog or rain reduce image contrast and clarity.
    (e) Aliasing artifacts caused by violation of the Nyquist sampling criterion, where high-frequency image components are inadequately sampled, leading to structural distortion and false patterns.
    (f) Low-light degradation at the sensor level, where reduced photon counts in dark scenes lower the signal-to-noise ratio and color accuracy.
    }
    \label{fig:degra}
\end{figure*}
As illustrated in Fig. \hyperref[fig:degra]{5}, the considered degradation types include LR, motion blur, defocus blur, weather degradation, aliasing, and low-light conditions. These factors can disrupt feature detection, camera pose estimation, and multi-view consistency, which ultimately reduces reconstruction quality. In response, various LLV methods have been proposed, including SR, deblurring, weather degradation removal, restoration, and enhancement.

\begin{figure*}[t]
    \centering
    \includegraphics[width=\textwidth]{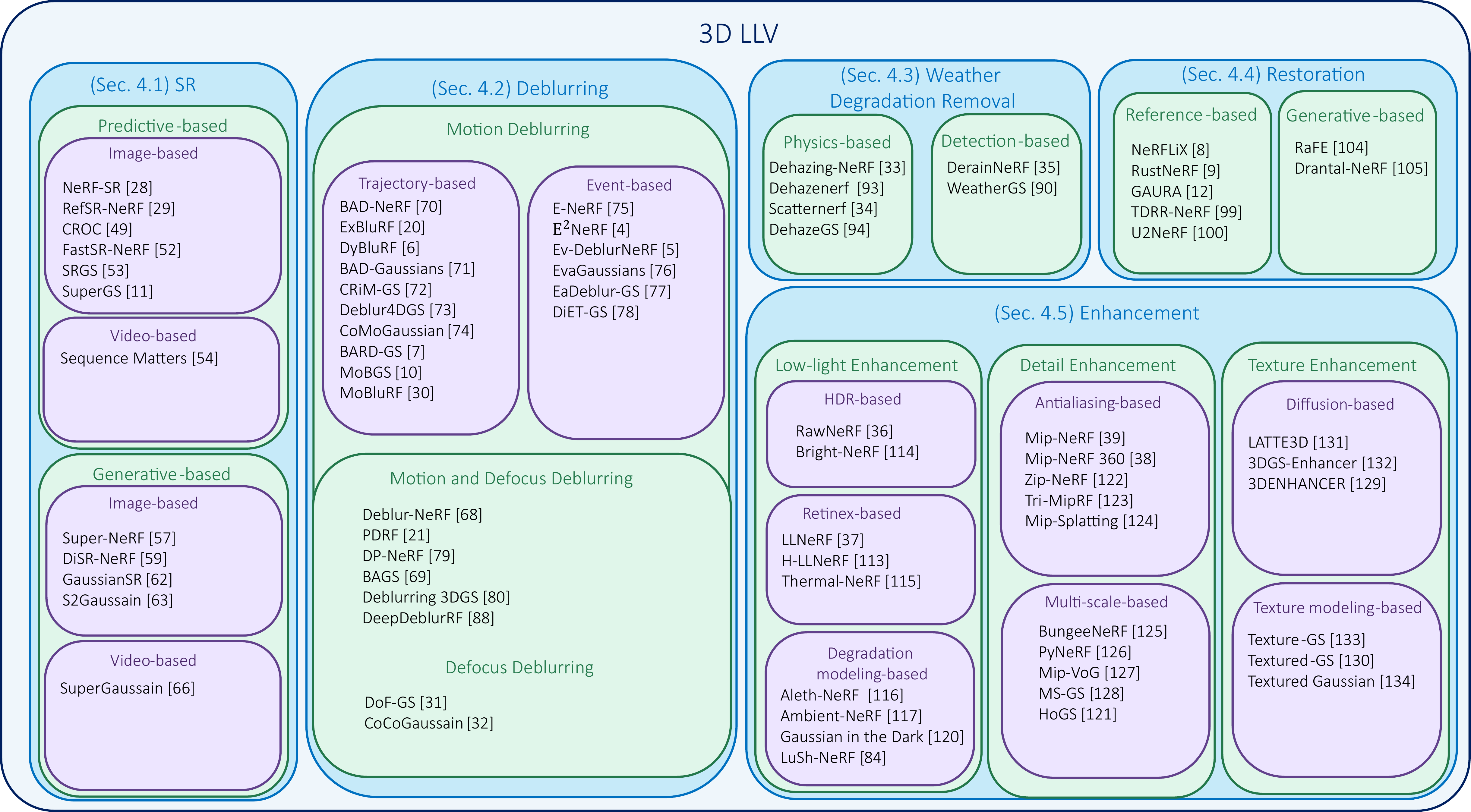}
    \caption{Taxonomy of 3D LLV tasks, including SR, Deblurring, Weather degradation removal, Restoration and Enhancement. Each task is categorized into various sub-methods based on its problem-solving strategies.}
    \label{fig:tax}
\end{figure*}
To address the aforementioned degradations, various enhancement strategies have been developed. As illustrated in Fig. \hyperref[fig:tax]{6}, these approaches can be broadly categorized into SR, deblurring, weather degradation removal, general enhancement, and restoration. The following sections describe the characteristics and roles of each method in improving 3D neural rendering performance under degraded conditions.

\textbf{SR} addresses the issue of structural information loss caused by LR images. In real-world scenarios, such images arise from limitations in sensor resolution, shooting distance, quality deterioration during compression or transmission, and other factors \cite{b5,b6}. Since these images lack clear object contours and fine structural details, they hinder the precise position estimation and view-consistency required by NeRF or 3DGS. As a result, accurate 3D reconstruction may be impossible, or blurred and geometrically distorted outputs may be generated.

\textbf{Deblurring} deals with the loss of high-frequency information due to motion blur and defocus blur in images. Motion blur occurs when light accumulates during exposure due to movement of the camera or objects \cite{b70}. Defocus blur, on the other hand, arises when a subject lies outside the camera's depth-of-field (DOF), causing rays from that region to diverge on the sensor \cite{b83,b84}. These blurs obscure scene boundaries and structural outlines, reducing the accuracy of spatial structure estimation. The presence of blur also weakens the correspondence between views, thereby degrading the performance of NeRF or 3DGS reconstruction.

\textbf{Weather degradation removal} addresses quality degradation caused by environmental factors such as haze, rain, or snow. Under such conditions, light undergoes scattering and absorption along the path between the camera lens and water particles suspended in the atmosphere, leading to reduced image contrast, blurred object contours, and inconsistencies in illumination and color \cite{b41,b43,b62}. This type of degradation introduces non-linear distortions in light propagation, making it difficult to maintain the view consistency that neural rendering frameworks rely on.

\textbf{Enhancement} addresses various forms of visual degradation that lead to information loss, such as low-light conditions, sensor noise, aliasing, and texture ambiguity. In low-light environments, the signal-to-noise ratio (SNR) decreases, resulting in distortion of brightness, color, and texture, and the suppression of fine structural details \cite{b115,b116}. Sensor-induced noise and compression artifacts further attenuate high-frequency components essential for spatial reconstruction. Aliasing artifacts cause discontinuities along object boundaries and disrupt view consistency, while weak or untextured regions hinder reliable feature extraction and correspondence matching \cite{b50,b53}. These factors serve as major sources of error in 3D reconstruction, degrading radiance parameter estimation and spatiotemporal consistency.

\textbf{Restoration} addresses cases where part of the information is missing from an image due to inter-view data loss, occlusions, sensor failures, or frame dropouts \cite{b99,b100}. Such missing information causes consistency issues in view-continuity-based neural rendering models, and unrecovered missing regions distort the overall geometric structure of the scene. Therefore, accurate 3D structure estimation is infeasible unless missing data are effectively recovered using surrounding information or multi-view interpolation mechanisms.

Considering the degradation types discussed above, including blur, LR, low-light conditions, and weather degradation, real-world visual inputs often exhibit a combination of these effects \hyperref[ref190]{[190]}, which severely degrade the performance of neural rendering frameworks such as NeRF and 3DGS that rely on consistent and HQ observations. To address these challenges, this paper systematically categorizes the types of degradation commonly encountered in real-world scenarios and provides a comprehensive survey of recent research that integrates LLV modules into 3D neural rendering pipelines. We emphasize the emerging role of 3D LLV as a crucial enabler of robust 3D reconstruction under adverse conditions, focusing on the methodological limitations and representative solutions designed to improve reconstruction quality and resilience.

%% file: sections/Preliminaries.tex
\section{Preliminaries} \label{PRELIMINARIES}
\subsection{Neural Radiance Fields (NeRF)}

NeRF \cite{b1} is an implicit radiance field representation that models a continuous 3D scene as a function. A multilayer perceptron (MLP) takes as input a 3D spatial location $x = (x, y, z)$ and a viewing direction $d = (\theta, \phi)$, where $\theta$ and $\phi$ denote the azimuth azimuth and elevation angles, respectively. The network outputs an RGB color $c = (r, g, b)$ and a volume density $\sigma$, formulated as:

\begin{equation}
F_\Theta(\mathbf{x}, \mathbf{d}) = (\mathbf{c}, \sigma),
\end{equation}
where $\Theta$ denotes the learnable parameters of the MLP.

Rendering is performed using ray-based volume rendering~\cite{b44, b45}. 
For each camera ray defined as $r(t) = \mathbf{o} + t\mathbf{d}$, where $\mathbf{o}$ is the ray origin and $\mathbf{d}$ is the viewing direction, and $t \in [t_n, t_f]$ defines the near and far bounds of the ray, the expected pixel color is computed by integrating the accumulated color contributions along the ray trajectory. 
The final color $\hat{C}(r)$ is defined by the following continuous volume rendering integral:

\begin{equation}
\hat{C}(\mathbf{r}) = \int_{t_n}^{t_f} T(t) \cdot \sigma(\mathbf{r}(t)) \cdot \mathbf{c}(\mathbf{r}(t), \mathbf{d}) \, dt,
\end{equation}

where $c(r(t), \mathbf{d})$ denotes the emitted RGB color at position $r(t)$ in direction $\mathbf{d}$, $\sigma(r(t))$ is the volume density, and $T(t)$ is the accumulated transmittance defined as:

\begin{equation}
T(t) = \exp\left(- \int_{t_n}^{t} \sigma(\mathbf{r}(s)) \, ds \right).
\end{equation}

In practice, this integral is approximated by discrete sampling along the ray \cite{b45}:

\begin{equation}
\hat{C}(\mathbf{r}) \approx \sum_{i=1}^{N} T_i \alpha_i \mathbf{c}_i,
\end{equation}

where $\alpha_i = 1 - \exp(-\sigma_i \delta_i)$, $T_i = \exp\left(- \sum_{j=1}^{i-1} \sigma_j \delta_j \right)$, and $\delta_i = t_{i+1} - t_i$ is the interval between adjacent sample depths $t_i$ and $t_{i+1}$.
Here, $N$ is the total number of discrete samples along the ray, and $i$ denotes the index of each sample.

To enable the representation of high-frequency scene details, NeRF adopts positional encoding, which transforms input coordinates into a higher-dimensional space using sinusoidal functions \cite{b93}. This alleviates the spectral bias of MLPs and enables learning of complex scene structures \cite{b8,b9}. Additionally, NeRF utilizes a coarse-to-fine hierarchical sampling strategy, where a coarse network estimates geometry to guide the sampling positions for a finer network.

The MLP parameters $\theta$ are optimized by minimizing reconstruction losses, such as the L1 distance between rendered and GT images. 

\subsection{3D Gaussian Splatting (3DGS)}
3DGS~\cite{b2} is an explicit radiance field representation method that models a scene using a large number of anisotropic 3D Gaussian primitives. Each primitive, indexed by $i$, is parameterized by a center position $\boldsymbol{\mu}_i \in \mathbb{R}^3$, a full 3D covariance matrix $\boldsymbol{\Sigma}_i \in \mathbb{R}^{3 \times 3}$, an opacity value $o_i \in [0,1]$, and spherical harmonics (SH)~\cite{b95,b111} coefficients $\boldsymbol{C}_i \in \mathbb{R}^k$ that represent view-dependent appearance. Here, $k$ denotes the degrees of freedom in SH and controls the angular frequency complexity of the view-dependent color function.

The covariance matrix $\boldsymbol{\Sigma}_i$ is reparameterized as:
\begin{equation}
\boldsymbol{\Sigma}_i = \boldsymbol{R} \boldsymbol{S} \boldsymbol{S}^\top \boldsymbol{R}^\top,
\end{equation}
where $\boldsymbol{S} = \mathrm{diag}(\boldsymbol{s}_i)$ is a diagonal matrix consisting of spatial scales $\boldsymbol{s}_i \in \mathbb{R}^3$, and $\boldsymbol{R} \in \mathrm{SO}(3)$ is a rotation matrix derived from a learnable quaternion.

Each primitive defines a Gaussian-shaped density distribution in 3D space. Given a spatial coordinate $\mathbf{X} \in \mathbb{R}^3$ within the scene volume, the Gaussian response is computed as:
\begin{equation}
G_i(\mathbf{X}) = \exp\left( -\frac{1}{2} (\mathbf{X} - \boldsymbol{\mu}_i)^\top \boldsymbol{\Sigma}_i^{-1} (\mathbf{X} - \boldsymbol{\mu}_i) \right).
\end{equation}

During rendering, all Gaussian primitives are projected onto the 2D image plane via splatting. Let $\mathbf{x}' \in \mathbb{R}^2$ denote the 2D projected location of a primitive center. For each pixel, the projected Gaussians are sorted by depth, and the final color is computed using $\alpha$-blending:
\begin{equation}
\mathbf{C} = \sum_{i \in \mathcal{N}_{\mathrm{cov}}} \mathbf{c}_i \alpha_i \prod_{j=1}^{i-1} (1 - \alpha_j),
\end{equation}
where $\mathcal{N}_{\mathrm{cov}}$ denotes the set of Gaussians covering a given pixel, $\mathbf{c}_i$ denotes the view-dependent RGB color decoded from the SH coefficients $\boldsymbol{C}_i$ under the current viewing direction, and $\alpha_i = o_i \cdot G_i(\mathbf{x}')$ is the opacity at the projected location $\mathbf{x}'$.

3DGS enables real-time rendering through a tile-based rasterizer~\cite{b94}, avoiding point sampling and achieving both high computational efficiency and rich spatial expressivity while maintaining the same alpha-compositing-based image formation model as NeRF.

Gaussian primitives are typically initialized using the sparse point clouds automatically generated during the SfM~\cite{b14,b15,b16,b17,b18} process with calibrated cameras.

Their attributes are optimized via backpropagation using a combination of L1 and SSIM losses~\cite{b57}. To improve reconstruction quality, 3DGS employs an adaptive densification process: primitives in under-reconstructed areas are cloned or split based on position gradients, while those in over-reconstructed or low-opacity regions are removed. This dynamic adjustment facilitates efficient and accurate modeling of high-resolution 3D scenes.

Furthermore, 3D Gaussians serve as a differentiable volumetric representation, allowing for accurate gradient computation during training. By tracking the traversal of sorted splats, the method supports visibility-aware anisotropic splatting and enables fast and stable backward passes.

%% file: sections/Low-Level_Vision_for_Robust_3D_Rendering.tex
\section{Low-Level Vision for Robust 3D Rendering} \label{Low-Level}
This section introduces representative 3D LLV methods for addressing various degradation conditions. The methods are presented in the order of SR, deblurring, weather degradation removal, restoration, and enhancement. These methods are organized as shown in Fig. \hyperref[fig:tax]{6}.

\subsection{Super-Resolution (SR) in 3D LLV}

3D SR aims to recover HR 3D representations from various 2D LR sources, including multi-view images or videos. Unlike traditional 2D SR, which focuses solely on spatial 2D pixel enhancement, 3D SR must account not only for spatial and temporal coherence but also for 3D geometric consistency across multiple views.

This paper categorizes 3D SR approaches according to their learning paradigms into predictive and generative methods. Each category is further divided based on input modality into image-based and video-based approaches. This taxonomy aligns with the structure of this section, and the overall classification is summarized in Fig. \hyperref[fig:tax]{6}.

\begin{figure}[t]
    \centering
    \includegraphics[width=\linewidth]{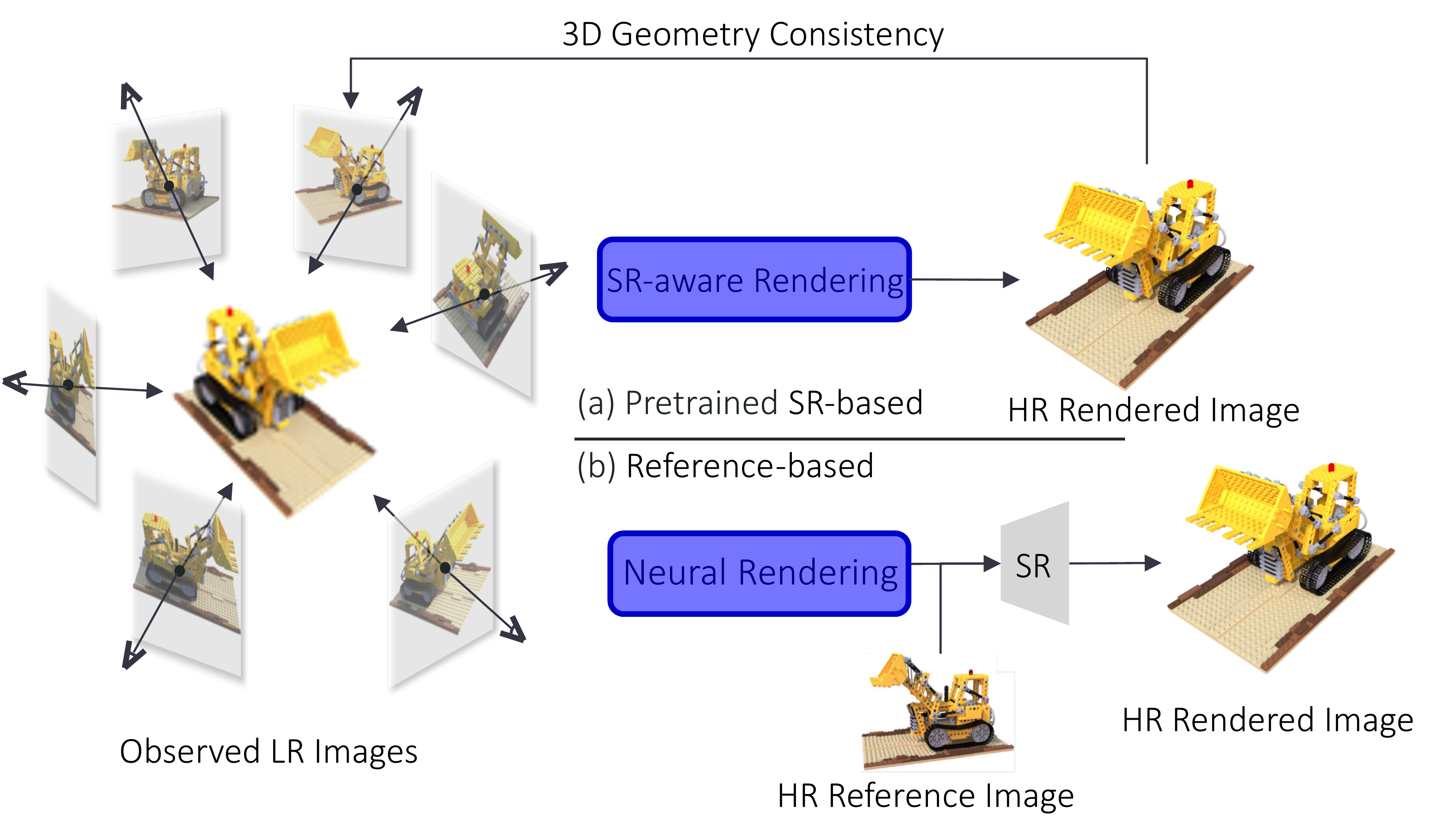}
    \caption{ Overview of representative SR integration strategies for 3D reconstruction or novel view synthesis.
    (a) Pretrained SR-based approaches apply a pretrained SR model \cite{b10,b11,b20, b33, b112} directly to the observed LR multi-view images, enabling SR-aware rendering with 3D geometric consistency.
    (b) Reference-based approaches utilize an additional HR reference image to guide neural rendering, followed by SR enhancement on the rendered output.
    }
    \label{fig:SR}
\end{figure}

Within the predictive paradigm, two representative strategies have emerged depending on how the SR module is integrated into the pipeline. Fig. \hyperref[fig:SR]{7} illustrates a structural comparison between pretrained SR-based and reference-based approaches.

\subsubsection{ Predictive Method}
Predictive approaches for 3D SR aim to enhance the resolution of multi-view inputs by leveraging Single-Image Super-Resolution (SISR) techniques or reference-based SR methods. These approaches are categorized into image-based and video-based frameworks, and are explicitly designed to preserve multi-view 3D geometric consistency, which is often disrupted when 2D SR is independently applied to each view. Their primary objective is to recover high-frequency details while maintaining multi-view consistency.

\paragraph{Image-based} 
NeRF-SR \cite{b5} introduces a supersampling strategy where each pixel is subdivided into sub-pixels, and multiple rays are cast per sub-pixel to generate HR supervision. Depth-based warping and patch refinement networks propagate localized high-frequency details from reference images to synthesized views.

RefSR-NeRF \cite{b6} proposes a two-stage architecture where a LR NeRF models low-frequency content and a CNN module leverages HR reference images to reconstruct high-frequency details. The joint optimization is performed via separate supervision signals for NeRF and CNN outputs.

CROC \cite{b7} combines a NeRF and a pre-trained SISR network \cite{b10, b11} through Cross-Guided Optimization. The SR Update Module fuses features from both components, enabling joint learning of 3D geometric consistency and texture fidelity.

FastSR-NeRF \cite{b172} introduces a lightweight NeRF and SR pipeline that improves rendering efficiency by combining a compact NeRF backbone with a pretrained CNN-based SR module \cite{b10}. Without requiring architectural modifications or additional supervision, the method enhances rendering speed by up to 18× while maintaining visual quality. A key contribution is random patch sampling, which serves as a simple yet effective augmentation strategy to improve SR generalization under limited training budgets

SRGS \cite{b12} adopts a supersplatting strategy where Gaussian primitives are subdivided in HR space, increasing sampling density. It integrates a pre-trained SISR model \cite{b11} to guide texture reconstruction without explicit HR supervision, while enforcing sub-pixel consistency for multi-view alignment.

SuperGS \cite{b13} extends 3DGS to HR rendering using a coarse-to-fine training strategy. A multi-resolution hash grid \cite{b111} is used to initialize a latent feature field, and a pre-trained SISR model \cite{b11} provides high-frequency guidance. Variational residual features are injected at the primitive level, and multi-view joint learning is applied to maintain spatial consistency across views.

\paragraph{Video-based}
Beyond image-based approaches, recent methods have expanded towards utilizing Video Super-Resolution (VSR) frameworks to further enhance HRNVS.  
Instead of relying on actual video sequences, these methods synthetically construct video-like sequences from unordered LR multi-view images and apply VSR models to the generated sequences.

Sequence Matters ~\cite{b28} constructs pseudo-video sequences from unordered multi-view LR images by leveraging both visual and geometric similarity. Visual similarity is computed using ORB \cite{b29}, a lightweight feature detector and descriptor widely used for efficient keypoint matching. The resulting sequences are fed into a VSR model \cite{b30} to generate HR outputs, enabling video-based SR without requiring actual video footage.

\subsubsection{Generative Method}

Generative approaches aim to synthesize high-frequency details directly from LR inputs without relying on external HR references, supporting HRNVS even under sparse or degraded input settings.
\paragraph{Image-based} 
Super-NeRF \cite{b19} generates HR novel views by exploring view-dependent latent vectors within the feature space of a pre-trained 2D SR model \cite{b20}. A pre-trained LR NeRF is used in early training, and a latent alignment step ensures consistency with the LR inputs.

DiSR-NeRF \cite{b21} alternates between NeRF rendering and 2D diffusion-based SR \cite{b112} in an Iterative 3D Synchronization (I3DS) framework. To enhance structural fidelity, it introduces Renoised Score Distillation (RSD), a hybrid strategy that combines Ancestral Sampling, which generates sharp yet inconsistent details, and Score Distillation Sampling (SDS) \cite{b23}, which produces smooth but geometrically stable outputs. RSD optimizes intermediate diffusion latents to achieve both high-frequency restoration and multi-view consistency.

GaussianSR \cite{b25} injects diffusion-based 2D SR \cite{b20} priors directly into 3D Gaussian primitives to enhance high-frequency details. It employs SDS for supervision, but the stochastic nature of diffusion can introduce structural artifacts. To address this, GaussianSR progressively narrows the sampling range during training and removes unstable Gaussians to improve 3D geometric consistency. This enables accurate HR reconstruction while preserving the efficiency of the 3DGS framework.

S2Gaussian \cite{b31} targets sparse-view scenarios by constructing a LR Gaussian representation from predicted depth maps \cite{b32}. It introduces Gaussian Shuffle Split, which increases sampling density by dividing a large Gaussian into multiple smaller Gaussians while preserving the original distribution’s mass and spatial variance. These primitives are then used to render pseudo-views, which are enhanced by a pre-trained 2D SR model \cite{b33} to supervise the final HR Gaussian optimization. This design allows accurate and consistent 3D reconstruction under limited view inputs.

\paragraph{Video-based}
As previously described, generative methods similarly adopt the strategy of constructing video-like sequences from unordered multi-view images and applying VSR models to enhance reconstruction quality.
SuperGaussian \cite{b26} first trains a 3DGS representation using LR images, then renders LR video sequences along manually defined camera trajectories. These sequences are used to fine-tune a pre-trained 2D VSR model \cite{b27}, and the resulting HR videos are subsequently used to re-optimize the 3DGS. This iterative process enables the generation of HR scene representations with both temporal and spatial consistency.
\subsection{Deblurring in 3D LLV}

3D deblurring aims to reconstruct sharp 3D scene representations from blurry multi-view images or videos. Unlike 2D deblurring, it must account for both 3D geometric consistency across viewpoints and temporal continuity across frames \cite{b65, b90}. In particular, camera pose estimation tools based on Structure-from-Motion (SfM), such as COLMAP \cite{b14}, often fail on severely blurred inputs due to the lack of reliable keypoints and correspondences \cite{b78}, limiting the applicability of neural rendering techniques like NeRF and 3DGS in real-world scenarios.

To address these challenges, recent approaches have been proposed. For motion blur, trajectory-based methods that model or estimate continuous camera and object motion \cite{b68, b69, b70, b71, b72, b73, b74, b75}, and event-based methods leveraging the high temporal resolution of event cameras \cite{b77, b78, b79, b80, b82, b85}, have shown promising results. For defocus blur, solutions include model-based methods that approximate optical blur using depth and aperture information \cite{b83, b84}, as well as kernel-based methods that restore sharpness using spatially varying blur kernels \cite{b65, b89, b90, b91}. Based on the type of blur and modeling approach, recent studies can be categorized into three groups: motion deblurring, defocus deblurring, and motion-and-defocus deblurring. This taxonomy is summarized in Fig. \hyperref[fig:tax]{6} which also reflects the structure of this section.

Neural rendering methods rely on the assumption of temporally and geometrically consistent input views and typically presume sharp images captured under negligible exposure time \cite{b68}. However, in real-world scenarios, rapid camera or object motion frequently introduces motion blur, violating these assumptions and substantially degrading reconstruction quality \cite{b65, b88, b89}. Motion blur distorts 3D spatial details and disrupts 3D multi-view consistency, leading to artifacts in the synthesized outputs \cite{b65}. Furthermore, accurate camera pose estimation is essential for training such models, yet motion-blurred images inherently encode integrated motion trajectories over the exposure duration, making precise pose recovery significantly more challenging \cite{b78}.
To mitigate these issues, recent works have incorporated motion-aware modeling into the rendering pipeline. Some approaches \cite{b77,b78,b79, b80, b82, b68, b69, b70} estimate continuous motion trajectories to temporally align blurred observations, while others explicitly disentangle camera and object motion to reconstruct dynamic scenes more faithfully \cite{b68}. In addition, event-based sensing has been explored to achieve temporally accurate deblurring.
These methods improve robustness under motion blur and enable reliable 3D reconstruction of dynamic scenes in realistic environments.

\subsubsection{Event-based} 

\begin{figure*}[t]
    \centering
    \includegraphics[width=0.9\linewidth]{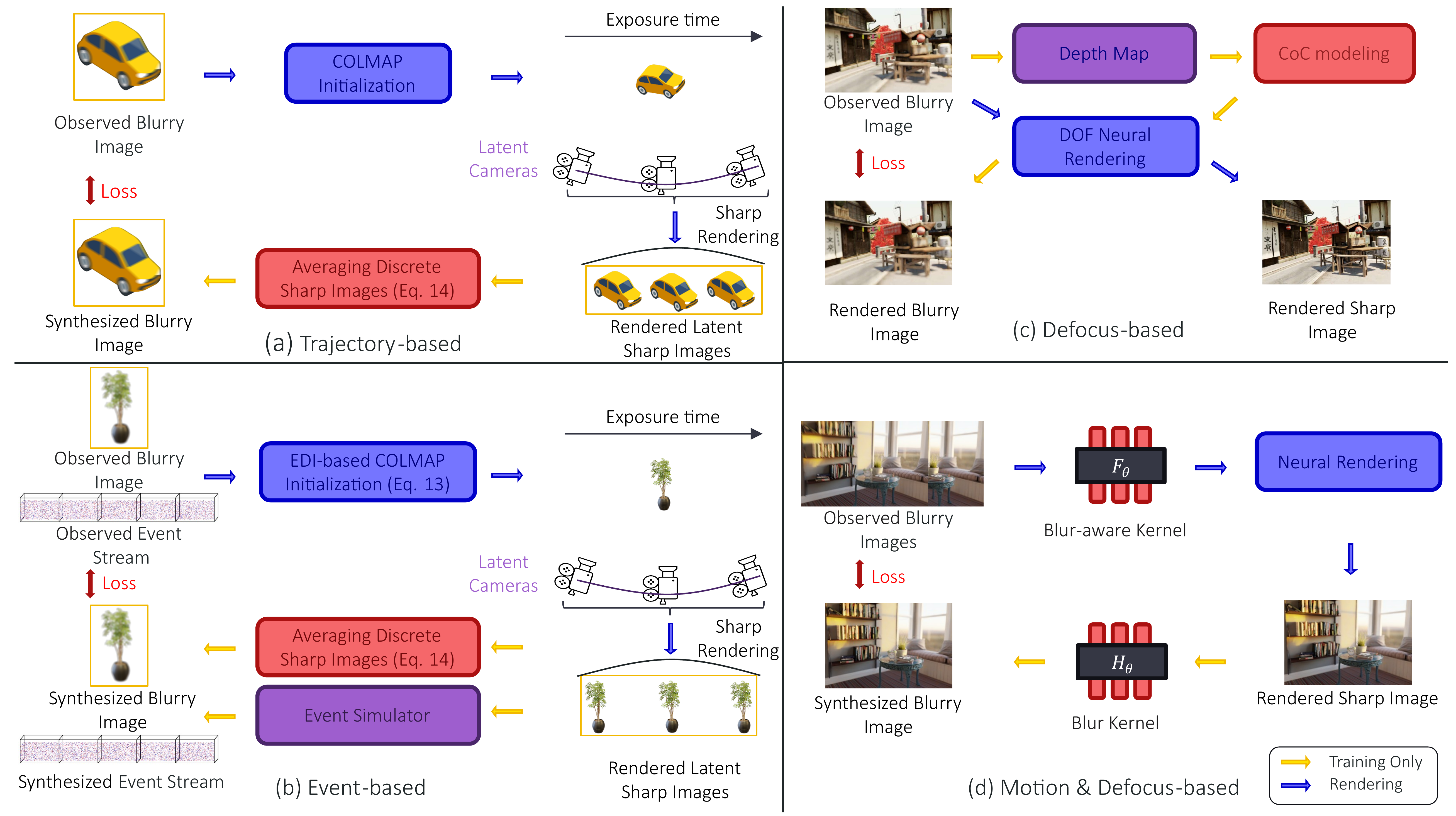}
    \caption{
    Four representative approaches for 3D deblurring. 
    (a) Trajectory-based methods initialize camera poses using COLMAP~\cite{b14, b15} and simulate the blur formation process by averaging sharp images rendered along a continuous camera trajectory. The synthesized blurry image is supervised by the observed input using a photometric loss. 
    (b) Event-based methods jointly leverage event streams and blurry images. The event simulator \cite{b178,b179} generates synthetic events, which are supervised by comparing them with ground-truth events, while the averaged sharp images are compared to the blurry image via a photometric loss. Camera poses are initialized using EDI-based COLMAP. 
    (c) Defocus-based methods employ depth maps and Circle-of-Confusion (CoC) modeling to enable differentiable DOF rendering. The rendered defocused image is supervised by the observed input. 
    (d) Motion and defocus-based methods jointly address motion blur and defocus blur by learning a blur-aware kernel \( F_\theta \) and a defocus blur kernel \( H_\theta \), which are used in a differentiable rendering pipeline. Yellow arrows indicate paths used only during training, while blue arrows denote rendering pipelines.
}
    \label{fig:deblur}
\end{figure*}

Event-based 3D deblurring leverages the unique characteristics of event cameras, which, instead of capturing full-frame intensity images at fixed intervals, produce a stream of asynchronous events that record per-pixel brightness changes. Each event is represented as $e = (x, y, t, p)$, where $(x, y)$ denotes the pixel coordinates, $t$ is the timestamp, and $p \in \{-1, +1\}$ indicates the polarity of the brightness change \cite{b77}. The overall process of event-based 3D deblurring is illustrated in Fig. \hyperref[fig:deblur]{8 (b)}.

Because the event stream is available at microsecond-level temporal resolution and exhibits a high dynamic range, event cameras are inherently robust to motion blur and low-light conditions. These advantages have been increasingly exploited in 3D deblurring frameworks, where the event stream serves as a supervisory signal for reconstructing sharp latent images, estimating camera motion trajectories, and constraining neural 3D representations such as NeRF and 3DGS under degraded input conditions.

E-NeRF \cite{b77} differs from conventional NeRFs in that it does not rely on direct comparisons with GT images. Instead, it employs an event generation-based supervision approach, where the predicted event-induced brightness changes are compared with those derived from real event streams to optimize motion blur.

$E^2$NeRF \cite{b78}, EvaGaussians \cite{b79}, EaDeblur-GS \cite{b80}, and DiET-GS \cite{b82} adopt the Event-based Double Integral (EDI) model \cite{b81} to robustly initialize camera poses in scenarios affected by strong motion blur. EDI leverages the physical relationship between asynchronous event streams and blurry RGB frames to reconstruct temporally sharp intensity images via temporal integration. This reconstruction is particularly advantageous for initializing SfM tools, which often fail on blurred inputs due to missing or mismatched feature correspondences.

Given a blurry image $B$, an exposure duration $\tau$, and a continuous event stream $E(t)$ recorded over the interval $[s - \tau/2, s + \tau/2]$, the EDI model formulates the image formation as:
\begin{equation}
B = I(s) \cdot \frac{1}{\tau} \int_{s - \tau/2}^{s + \tau/2} \exp(c \cdot E(t)) \, dt ,
\end{equation}
where $I(s)$ denotes the latent sharp image at the mid-exposure time $s$, and $c$ is the contrast threshold of the event camera. Sharp images at arbitrary time $t$ within the exposure window are estimated via:
\begin{equation}
I(t) = I(s) \cdot \exp(c \cdot E(t)) .
\end{equation}
The resulting set of sharp latent images $\left\{ I(t_i) \right\}_{i=1}^n$ warped to multiple uniformly sampled timestamps \cite{b79}, are then used for camera pose estimation and point cloud initialization with COLMAP. After initialization, these methods further refine the camera trajectories by simulating the formation of motion-blurred images. To do so, $n$ sharp images $\left\{ I(t_i) \right\}_{i=1}^n$, discretely sampled at uniform time intervals along the estimated continuous camera trajectory over the exposure time $\tau$, are rendered, and the resulting blurry image is obtained by temporal integration:
\begin{equation}\label{eq12}
\hat{B} = \frac{1}{n} \sum_{i=1}^{n} I(t_i) .
\end{equation}

This synthesized blurry image $\hat{B}$ is then compared with the real blurry input $B$ to supervise the joint optimization of the scene representation and the camera motion. 

In addition to the EDI-based supervision~\cite{b81}, these models also incorporate an event loss, which leverages the high temporal resolution of the event stream to constrain the brightness consistency between synthesized and observed events. While the specific formulations differ across works, the general principle involves comparing the brightness changes or event accumulation inferred from the rendered sharp latent images with the GT event data, thereby providing fine-grained supervision even under severe motion blur.

DiET-GS \cite{b82} builds on the EDI-based initialization and joint optimization framework by introducing cycle consistency regularization across supervisory signals and leveraging a generative prior from a pretrained diffusion model \cite{b122} to enhance the visual quality of rendered images.
Ev-DeblurNeRF \cite{b85} directly supervises NeRF outputs using brightness values derived from EDI-based latent sharp images and integrates temporally precise event stream information into the training process.

\subsubsection{Trajectory-based} 

In 3D scene reconstruction, camera or object motion during exposure is modeled as a continuous trajectory. By rendering sharp images along the estimated path and averaging them over time, the blur formation process is physically simulated \cite{b68, b69, b71, b72}. This approach aligns with the blur synthesis mechanism in Eq.~\ref{eq12}, enabling precise supervision and robust optimization under motion blur, while maintaining spatial and temporal consistency in dynamic scenes. The overall pipeline of trajectory-based deblurring is illustrated in Fig. \hyperref[fig:deblur]{8 (a)}.

Recent trajectory-based 3D deblurring methods, such as BAD-NeRF \cite{b68}, ExBluRF \cite{b69}, MoBluRF \cite{b70}, DyBluRF \cite{b71}, BAD-Gaussians \cite{b72}, CRiM-GS \cite{b73}, Deblur4DGS \cite{b74}, CoMoGaussian \cite{b75}, and BARD-GS \cite{b76}, commonly adopt a strategy that models the camera motion during exposure time as a continuous trajectory. They synthesize a blurry image by rendering multiple sharp images along the estimated trajectory and averaging them over time. These synthesized blurry images are then compared with the given blurry inputs to minimize photometric loss, enabling joint optimization of both the camera trajectory and the 3D scene representation.

Among them, BAD-NeRF, DyBluRF, BAD-Gaussians, and Deblur4DGS specifically define the camera trajectory using interpolation in the Lie group $\mathrm{SE}(3)$, where the trajectory is parameterized by only two learnable poses: the start pose $T_\text{start}$ and the end pose $T_\text{end}$. The intermediate pose at time $t \in [0, \tau]$ is computed using the following equation:

\begin{equation}
T_t = T_\text{start} \cdot \exp\left( \frac{t}{\tau} \cdot \log\left( T_\text{start}^{-1} \cdot T_\text{end} \right) \right) .
\end{equation}

Here, $\tau$ denotes the exposure time. This formulation allows the generation of continuous camera poses within the exposure interval by interpolating on the Lie algebra $\mathfrak{se}(3)$ and mapping back to $\mathrm{SE}(3)$ using the exponential map. The rendered images at sampled timestamps $t$ are then averaged to synthesize a motion-blurred image, and the entire pipeline is trained by minimizing the photometric difference between the synthesized and observed blurry images.

In addition, ExBluRF defines the camera trajectory not through interpolation in $\mathrm{SE}(3)$, but instead using a B\'ezier curve representation. This approach models the camera motion during the exposure time as a high-order curve defined by control points, allowing smoother and more flexible trajectory generation over time.
The B\'ezier curve is parameterized by time $\tau \in [0, 1]$ and defined using $M+1$ control points $\{\hat{\mathbf{p}}_j\}_{j=0}^{M}$ in the Lie algebra $\mathfrak{se}(3)$ \cite{b86}, as follows:

\begin{equation}
\mathbf{p}_{\tau} = \sum_{j=0}^{M} \binom{M}{j}(1 - \tau)^{M - j} \tau^j \cdot \hat{\mathbf{p}}_j . 
\end{equation}

Each control point $\hat{\mathbf{p}}_j$ lies in $\mathfrak{se}(3)$, and the corresponding camera pose $\mathbf{P}_{\tau}$ in $\mathrm{SE}(3)$ can be obtained via the exponential map. This formulation is particularly suitable for scenarios involving longer exposure durations or complex curved camera trajectories, as it naturally ensures smoothness and differentiability across the entire motion path.
CRiM-GS and CoMoGaussian adopt a neural ODEs \cite{b87} framework to model camera motion trajectories continuously over time. Unlike discrete interpolation methods such as $\mathrm{SE}(3)$ or B\'ezier curves, neural ODEs parameterize time-continuous transformations with high flexibility and differentiability.
Specifically, given an initial latent state $\mathbf{z}(\tau_0)$ that encodes the screw axis or rigid motion parameters, the continuous trajectory is defined by the solution of an ODE:
\begin{equation}
\frac{d\mathbf{z}(\tau)}{d\tau} = f(\mathbf{z}(\tau), \tau; \phi),
\end{equation}

where $f$ is a neural network parameterized by $\phi$. The latent state at any timestamp $\tau_s$ is then obtained via integration:

\begin{equation}
\mathbf{z}(\tau_s) = \mathbf{z}(\tau_0) + \int_{\tau_0}^{\tau_s} f(\mathbf{z}(\tau), \tau; \phi) \, d\tau.
\end{equation}

This continuous integration process enables the computation of camera poses at arbitrary timestamps within the exposure interval, allowing the construction of a dense and temporally smooth motion trajectory.

MoBluRF~\cite{b70}, DyBluRF~\cite{b71}, Deblur4DGS~\cite{b74}, BARD-GS~\cite{b76}, and MoBGS~\cite{b97} are video-based 3D deblurring methods that take monocular video as input and commonly adopt a strategy of separating static and dynamic regions during training. These approaches use masks \cite{b113} to identify dynamic object areas and optimize the static and dynamic parts independently, enabling effective learning of both camera motion and object motion.

In particular, MoBluRF \cite{b70} initializes each ray individually and then refines them progressively by splitting into multiple rays. Each ray is optimized to reflect continuous motion throughout the exposure time, allowing the model to represent the motion of the camera and the object separately. In contrast, DyBluRF \cite{b71}, Deblur4DGS \cite{b74}, and BARD-GS \cite{b76} explicitly learn camera trajectories and model motion blur based on continuous motion within the exposure interval. By incorporating motion-aware learning frameworks and trajectory-based supervision, these methods enable accurate 3D reconstruction even in video scenes with complex motion blur. Building upon these ideas, MoBGS~\cite{b97} introduces a physically-consistent blur formation model for 3DGS by jointly estimating the camera motion trajectory and the adaptive exposure time from blurry observations. This enables unsupervised deblurring and temporally coherent novel view synthesis in dynamic scenes, even under complex motion blur.

\subsubsection{Defocus Deblurring} 
Neural rendering methods generally assume that all scene points are in focus and rely on sharp, depth-consistent views for accurate reconstruction~\cite{b83}. However, in real-world imaging, limited DOF and large apertures often cause defocus blur, where out-of-focus points are projected as finite-sized blur circles, known as the Circle of Confusion (CoC)~\cite{b83,b84}. The CoC radius, derived from the thin-lens model, increases as the object distance deviates from the focus distance, leading to spatially varying blur:

\begin{equation}
R_{\text{CoC}} = \frac{1}{2} Q \left| \frac{1}{z_o} - \frac{1}{f} \right|, \quad Q = F \cdot A,
\end{equation}

where $z_o$ is the distance to the scene point, $f$ the focus distance, $F$ the focal length, and $A$ the aperture diameter. This spatially varying defocus blur, which increases with distance deviation from the focal plane, is illustrated in Fig. \hyperref[fig:deblur]{8 (c)}.

Such defocus blur disrupts geometric consistency and fine detail reconstruction, especially near depth discontinuities~\cite{b83}. To address this, recent works~\cite{b83,b84} introduce physically grounded defocus modeling into neural rendering pipelines. These approaches estimate CoC from depth and camera parameters, simulate depth-aware blur during training, and in some cases, optimize learnable focus and aperture settings for flexible post-capture refocusing.

DOF-GS~\cite{b83} introduces two learnable parameters per view: focal distance $f_m$ and aperture factor $Q_m$, where the subscript $m$ denotes the index of the view. A depth-aware blur kernel is applied to each 2D-projected Gaussian based on the computed CoC radius. An all-in-focus (AiF) image rendered with zero aperture is used to supervise sharp region localization via an auxiliary network.

CoCoGaussian~\cite{b84} explicitly constructs the CoC shape by placing multiple CoC Gaussians in a circular layout around each base Gaussian based on the CoC radius. Each CoC Gaussian is displaced along a unit direction vector, scaled by a learnable factor. The base and CoC Gaussians are rendered separately and combined through a weighted sum to generate the final defocused image.

\subsubsection{Deblurring for Motion and Defocus}  
In real-world imaging scenarios, motion blur and defocus blur frequently occur together due to fast camera movement and shallow DOF~\cite{b90}. Under such compounded blur conditions, the core assumptions of neural rendering methods, including the use of sharp inputs and the consistency of views across observations, are severely violated. This results in significant degradation in reconstruction quality~\cite{b88}. To overcome this challenge, recent approaches~\cite{b65, b88, b89, b90, b91, b92} have integrated blur modeling directly into the neural rendering pipeline. This integration enables robust training and accurate scene reconstruction even under realistic visual degradations. The overall framework for joint modeling of motion and defocus blur in neural rendering pipelines is illustrated in Fig. \hyperref[fig:deblur]{8 (d)}.

Deblur-NeRF \cite{b65} models blur by sampling rays from nearby positions and computing weighted combinations of their colors. This captures both motion and defocus effects and jointly optimizes ray offsets, weights, and NeRF.

PDRF \cite{b88} uses a coarse-to-fine strategy by first estimating a blur field and LR geometry, then refining with HR voxel representation. It ensures stable training under heavy blur conditions.

DP-NeRF \cite{b89} simulates blur by applying rigid-body transformations to rays in $\mathrm{SE}(3)$, accounting for motion and occlusion. It maintains spatial consistency across views during blur modeling.

BAGS \cite{b90} predicts per-pixel degradation masks and blur kernels using a 2D prior network, guiding the optimization of 3D Gaussians. It refines geometry and blur jointly for robust reconstruction.

Deblurring 3DGS \cite{b91} encodes blur effects by deforming Gaussian covariances without explicit kernels. It adapts the anisotropy and spread of each Gaussian during training to model motion and defocus.

DeepDeblurRF \cite{b92} combines a 2D image deblurring network \cite{b114} with radiance field learning. The enhanced 2D output supervises the 3D optimization process, enabling integration of pre-trained 2D restorers into 3D rendering.

\subsection{Weather Degradation Removal in 3D LLV}
NeRF and 3DGS are highly sensitive to the quality of images \cite{b68}, and adverse weather conditions such as rain, haze, and snow often disrupt multi-view consistency and hinder accurate structural learning \cite{b62, b63}. To address this challenge, recent studies have explored two main approaches for restoring image quality under weather-degraded conditions. The first approach is physics-based methods, which compensate for weather effects by mathematically modeling the underlying degradation factors. The second approach is detection-based methods, which identify weather-degraded regions and prevent loss from being applied to them during training, thereby allowing multi-view supervision to naturally recover the degraded content without direct correction.
This classification of weather degradation removal methods is summarized in Fig. \hyperref[fig:tax]{6}.

\begin{figure}[t]
    \centering
    \includegraphics[width=\linewidth]{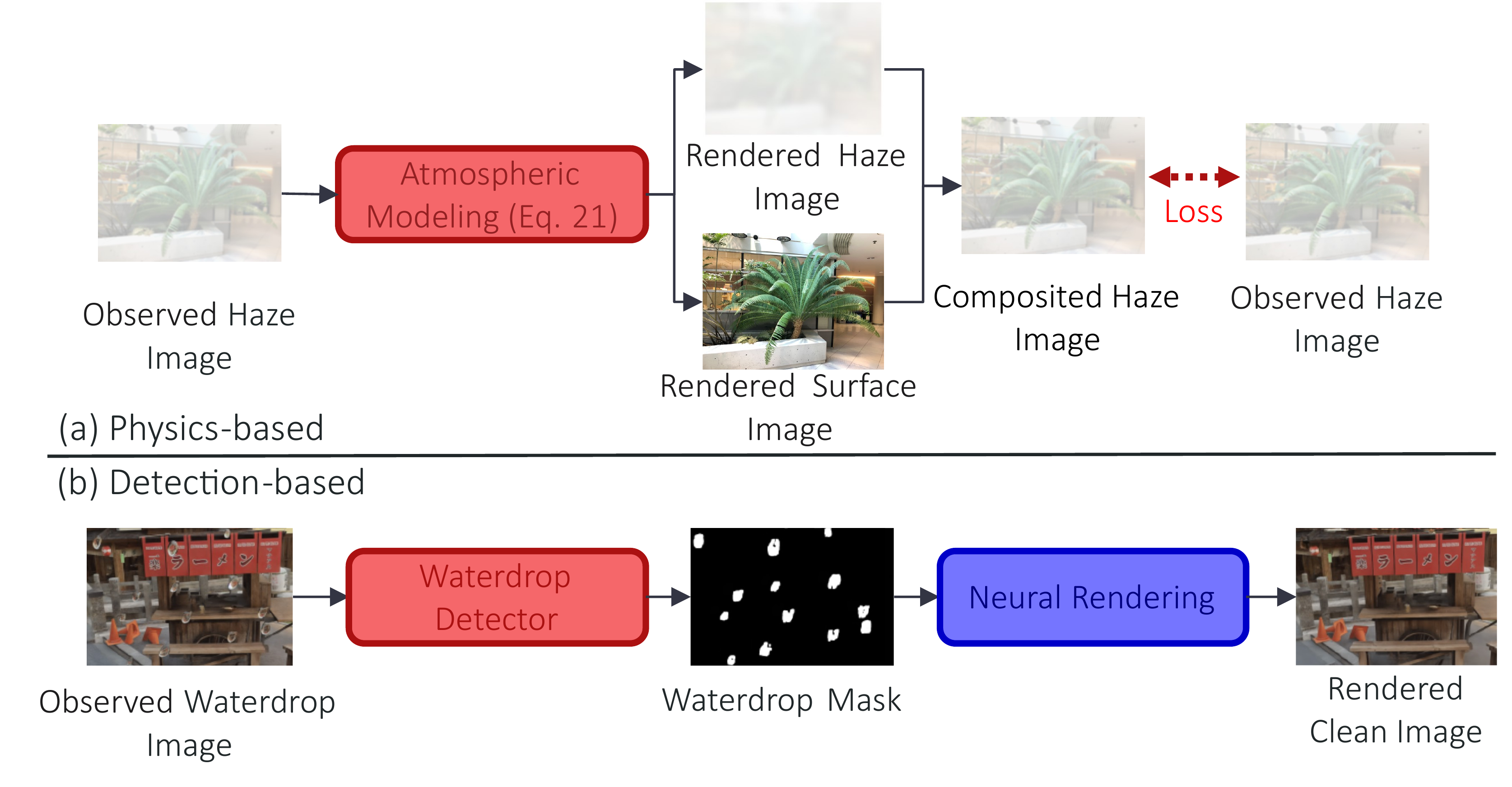}
    \caption{
    Two representative approaches for weather degradation removal. 
    (a) Physics-based methods rely on atmospheric light modeling to simulate scattering and absorption in the atmosphere. The observed degraded image is decomposed into a rendered surface and a rendered haze component, and the reconstructed haze image is composited with the rendered surface. Training is guided by a reconstruction loss between the composited image and the observed input. This approach explicitly separates degradation from surface appearance through physically grounded modeling \cite{b39}. 
    (b) Detection-based methods use a pretrained detector \cite{b64} to identify localized weather-induced degradation (e.g., waterdrops, mist, or dirt) and generate a binary mask. Neural rendering is then performed by excluding the masked regions from supervision, enabling robust reconstruction without relying on corrupted pixels.
}

    \label{fig:weather}
\end{figure}
 
\subsubsection{Physics-based} 

To address the degradation caused by atmospheric scattering in multi-view images, recent neural rendering methods integrate physics-based haze models into volumetric rendering frameworks \cite{b34,b41,b43,b46}. These approaches aim to disentangle the effects of haze from scene radiance while enabling geometrically consistent and photorealistic 3D reconstructions under adverse weather conditions, as shown in Fig. \hyperref[fig:weather]{9 (a)}. Depending on the underlying physical model, these methods can be broadly divided into two categories. The first group is based on the Koschmieder model \cite{b39}, and the second adopts the Atmospheric Scattering Model (ASM). For each group, we describe the shared assumptions and core formulations and then discuss the main contributions of each representative method.

\paragraph{Koschmieder-based}

DehazeNeRF~\cite{b34} and ScatterNeRF~\cite{b43} are representative neural rendering methods that adopt the Koschmieder model \cite{b39} to simulate haze formation. Both methods assume that the observed radiance is a combination of exponentially attenuated scene radiance and additive airlight. This classical model is defined as

\begin{equation}
    C(x) = J(x) \cdot e^{-\beta d(x)} + A \cdot (1 - e^{-\beta d(x)}),
\end{equation}

where $C(x)$ is the observed hazy radiance, $J(x)$ is the haze-free radiance, $\beta$ is the scattering coefficient, $A$ is the global airlight, and $d(x)$ is the scene depth. These approaches extend the original formulation to volumetric rendering by separating scene and haze densities and applying depth-aware integration along rays. Regularization strategies are introduced to ensure physical plausibility and disentanglement.

DehazeNeRF generalizes the Koschmieder model by incorporating a physically grounded rendering formulation based on the Radiative Transfer Equation \cite{b35}. It decomposes the accumulated radiance into surface and haze components as

\begin{equation}
\begin{split}
C(r, d) =\ & \underbrace{\int_{t_n}^{t_0} c(r(t), d)\, \sigma(t)\, T_{\sigma+\sigma_s}(t)\, dt}_{\text{Surface}} \\
          & + \underbrace{\int_{t_n}^{t_0} c_s(r(t))\, \sigma_s(t)\, T_{\sigma+\sigma_s}(t)\, dt}_{\text{Haze}},
\end{split}
\end{equation}

where $\{c, \sigma\}$ and $\{c_s, \sigma_s\}$ represent the surface and haze properties respectively. Surface opacity is defined using a Signed Distance Function (SDF), and haze parameters are modeled with a low-frequency MLP. During training, Koschmieder consistency and the Dark Channel Prior (DCP) \cite{b40} are employed as regularizers. At inference time, the haze components are removed to render a clear scene.

ScatterNeRF simplifies the modeling by adhering strictly to the Koschmieder law and explicitly disentangles haze and scene volumes. It models the observed color as a convex combination of clear radiance and airlight using

\begin{equation}
    C_F = l \cdot C_c + (1 - l) \cdot C_p, \quad l = \exp(-\sigma_p D),
\end{equation}

where $C_c$ and $C_p$ are the radiance from the clear scene and airlight, $\sigma_p$ is the fog density, and $D$ is the depth. Each component is predicted by a separate MLP. The method encourages volumetric separation through entropy-based regularization without relying on handcrafted priors such as DCP. This formulation allows for flexible control over haze simulation and removal.

\paragraph{ASM-based}
Dehazing-NeRF~\cite{b41} and DehazeGS~\cite{b46} are representative methods that adopt the Atmospheric Scattering Model (ASM) \cite{b42} to simulate haze formation and removal in neural rendering pipelines. While similar in form to the Koschmieder model, ASM-based methods typically integrate the haze formation model in a modular way, combining atmospheric parameter estimation with learned scene geometry or radiance. The general image formation equation is given as

\begin{equation}
I(x) =\ J(x) \cdot e^{-\beta d(x)}  + A \cdot \left(1 - e^{-\beta d(x)}\right),
\end{equation}

where $I(x)$ is the observed hazy image, $J(x)$ is the clean scene radiance, $\beta$ is the scattering coefficient, $A$ is the global airlight, and $d(x)$ is the depth of the scene. Both methods incorporate this model into the rendering pipeline to enable joint learning or estimation of haze-free content and atmospheric parameters.

Dehazing-NeRF proposes a dual-branch architecture that jointly estimates ASM parameters and reconstructs a clean NeRF representation. One branch utilizes a pretrained network to estimate $\beta$ and $A$, while the other learns a clean radiance field via volume rendering. The predicted depth and radiance are then fused through the ASM to reconstruct hazy images. To improve robustness and avoid convergence to trivial solutions, the method introduces soft-margin reconstruction consistency, atmospheric consistency loss, and contrast discriminative loss. 

DehazeGS integrates the ASM into the 3DGS framework by predicting transmittance values for each Gaussian. These are inferred using a one-dimensional convolutional network that maps normalized Gaussian depth to transmittance, with gradients detached to preserve disentanglement. The final hazy radiance is rendered by blending the latent clear Gaussian representation with a learnable global airlight using the predicted transmittance. Pseudo-depth supervision is used to guide training, and regularization is applied through the DCP and Bright Channel Prior (BCP) \cite{b47}.

\subsubsection{Detection-based} 
Detection-based methods are designed to address localized degradations, such as adherent raindrops, snowflakes, and lens occlusions. Unlike volumetric degradations caused by atmospheric scattering, these artifacts are concentrated in specific regions of the image and serve as a major source of multi-view inconsistencies, ultimately hindering accurate 3D structure learning \cite{b62}.

To mitigate these challenges, both DerainNeRF \cite{b62} and WeatherGS \cite{b63} adopt a masking-based strategy that explicitly identifies degraded regions and excludes them from the training process. By leveraging pretrained detection modules, each method generates binary masks that indicate degraded areas in the input images, ensuring that corrupted observations are removed from the optimization pipeline. This selective masking prevents error propagation from degraded pixels and allows the network to focus on learning clean scene representations, as shown in Fig. \hyperref[fig:weather]{9 (b)}.

DerainNeRF employs a pretrained raindrop detector \cite{b64} to predict the locations of adherent waterdrops. Based on these predictions, binary masks are generated and used to exclude occluded pixels during NeRF training. To further improve mask accuracy, especially in cases where waterdrops adhere to the camera lens and remain consistent across views, attention maps from multiple frames are averaged to enhance detection robustness.

WeatherGS categorizes degradation into two types. High-density particles such as snow and rain are filtered out using an Atmospheric Effect Filter, while occlusions caused by precipitation on the lens are detected through a Lens Effect Detector \cite{b64}. The resulting masks are then applied during 3DGS to ensure that occluded regions do not interfere with the learning of scene representations.
By explicitly detecting and excluding weather-induced degradations from the learning process, detection-based methods enable structurally accurate and visually consistent 3D scene reconstruction, even under complex and adverse weather conditions.

\subsection{Restoration in 3D LLV}
\begin{figure}[t]
    \centering
    \includegraphics[width=\linewidth]{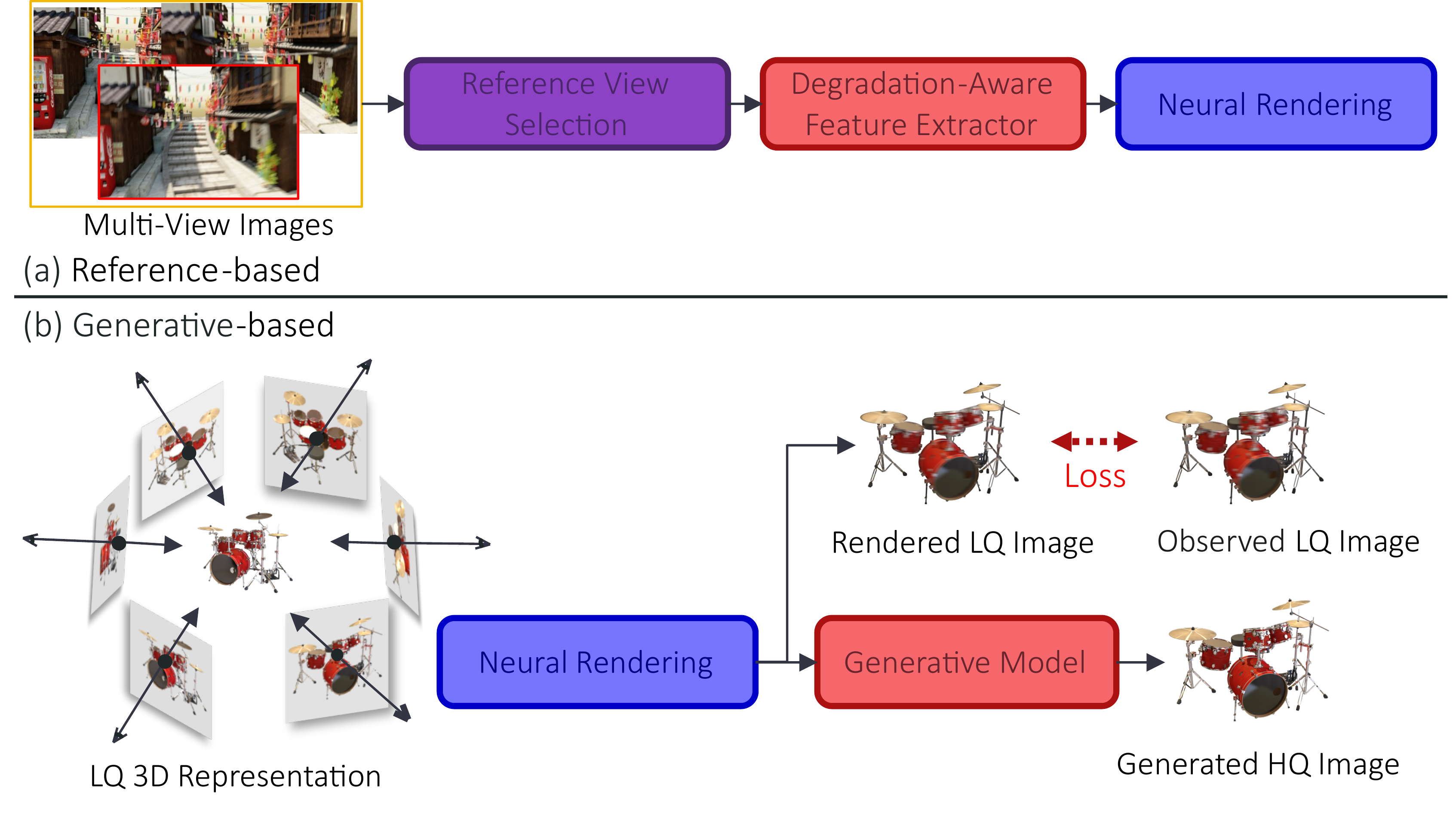}
    \caption{Overview of representative 3D restoration approaches in neural rendering.
    (a) Reference-based methods utilize neighboring views to restore degraded regions by leveraging spatial redundancy and cross-view consistency.
    (b) Generative model-based methods synthesize restored content directly from degraded observations using learned priors, enabling restoration without explicit references.}
    \label{fig:restoration}
\end{figure}
3D restoration aims to recover high-fidelity 3D scene representations from degraded 2D multi-view images or video sequences. In real-world scenarios, multi-view captures often suffer from various forms of degradation due to limitations of imaging devices, insufficient lighting, motion blur, defocus blur, compression artifacts, and sensor noise~\cite{b99}. Such degradations not only reduce the visual quality but also severely disrupt the geometric and temporal consistency across multiple views, making accurate 3D reconstruction challenging. In particular, SfM pipelines, which rely on reliable keypoint detection and correspondence matching~\cite{b78}, frequently fail under degraded conditions. As a result, neural rendering methods experience significant quality deterioration when applied to real-world degraded inputs.

To address these challenges, recent studies commonly adopt a strategy that synthetically simulates real-world degradations to create degraded multi-view datasets for supervised training. Major degradation simulation techniques include the addition of Gaussian noise, application of motion and defocus blur, insertion of compression artifacts, and simulation of low-light conditions~\cite{b99, b100}. Some methods further incorporate high-order degradation modeling or combine multiple degradation types to generate more realistic degraded data. Based on such synthetic degraded datasets, various restoration approaches have been proposed.

Restoration methods are typically categorized into reference-based and generative model-based approaches, each adopting distinct strategies for degradation handling and network design, as illustrated in Fig. \hyperref[fig:restoration]{10}

\subsubsection{Reference-based}
Reference-based restoration methods aim to recover high-quality target views from degraded inputs by leveraging multiple reference views with similar viewpoints. These approaches typically aggregate information from nearby views to maintain structural consistency and cross-view alignment. Representative methods include NeRFLiX~\cite{b99}, RustNeRF \cite{b98}, GAURA~\cite{b100}, TDRR-NeRF~\cite{b105}, and U2NeRF~\cite{b102}.

NeRFLiX~\cite{b99} selects two 2D reference views with the highest overlap with the target view and applies window-based attention at both patch and pixel levels to aggregate features. It employs a coarse-to-fine attention refinement strategy to progressively enhance the quality of high-frequency details.

RustNeRF~\cite{b98} explicitly models high-order degradations and restores multi-view degraded inputs by extracting features from several neighboring reference views, spatially aligning them, and fusing them via attention-based mechanisms. This redundancy-enhanced feature fusion helps maintain multi-view consistency and improves restoration performance.

GAURA~\cite{b100} extends the GNT framework to generate high-quality novel views from degraded multi-view inputs. It employs a Degradation-aware Latent Module inspired by HyperNetworks~\cite{b149} to produce transformation weights based on degradation-specific latent codes. Additionally, it extracts residual features from the reference view closest to the target to address intra-class variation. These degradation-aware transformations are applied during epipolar-based feature aggregation to ensure multi-view consistency.

TDRR-NeRF~\cite{b105} is a lightweight, model-agnostic feature module that enhances the robustness of Generalizable NeRF (GNeRF)~\cite{b150} to degraded inputs. The framework consists of two stages. First, a self-supervised depth estimator predicts a coarse depth map and aligns nearby degraded views to the source. Then, a 3D-aware feature extractor fuses the aligned features with the source view to correct distortions caused by degradation. An auxiliary restoration head further refines the features through residual learning, supervised by clean source images.

U2NeRF~\cite{b102} is a self-supervised framework that performs joint degradation restoration and novel view synthesis from underwater multi-view inputs. Built upon the Generalizable NeRF Transformer (GNT)~\cite{b106}, U2NeRF aggregates multi-view features along epipolar lines and integrates them along ray directions using stacked view and ray transformers. While conventional NeRF operates on a per-pixel basis, which is insufficient for underwater restoration, U2NeRF extends ray features to the 2D patch level. Underwater images are decomposed into several components: scene radiance $J$, direct transmission map $T_D$, backscatter transmission map $T_B$, and global background light $A$. The reconstructed image is defined as:
\begin{equation}
I(i) = J(i) T_D(i) + (1 - T_B(i)) A.
\end{equation}
Each component is predicted either from ray features or via a Variational Autoencoder (VAE) using the nearest reference view, enabling restoration without GT supervision.

\subsubsection{Generative-based}

Generative-based restoration employs pretrained generative models to recover degraded inputs. RaFE~\cite{b104} and Drantal-NeRF~\cite{b103} adopt different generative paradigms. Drantal-NeRF uses diffusion models and RaFE utilizes GANs incorporates world models to handle various types of degradation.

RaFE adopts a generative adversarial learning strategy to address multi-view inconsistencies that arise from individually restored views. In the first stage, each degraded view is independently restored using pretrained 2D restoration models \cite{b108,b109,b110}, and in the second stage, a StyleGAN2-like CNN-based generator\cite{b107}  adds fine-level residual features to a fixed coarse tri-plane. The generator is supervised using adversarial loss and perceptual loss to ensure visually realistic and structurally consistent rendered views. RaFE effectively reconstructs photorealistic and geometrically accurate 3D scenes under various degradations, including blur, noise, LR, and mixed artifacts.
\begin{figure}[t]
    \centering
    \includegraphics[width=\linewidth]{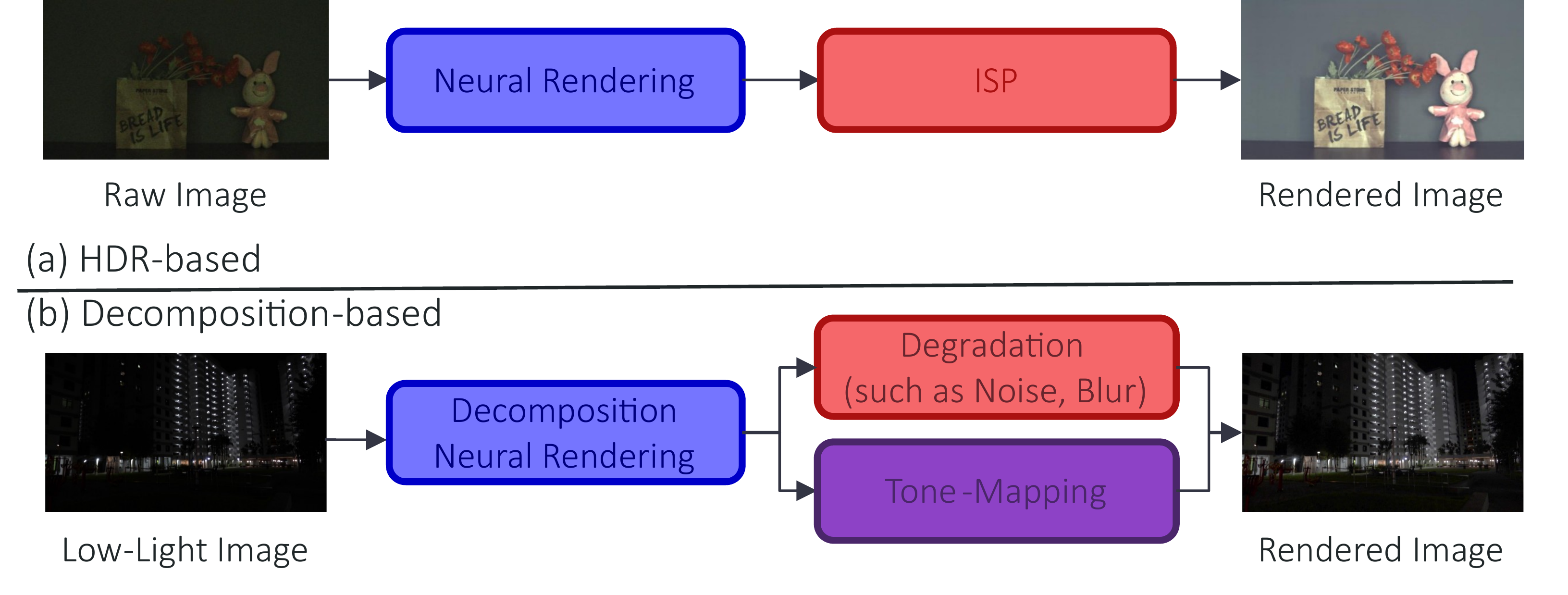}
    \caption{
    Two representative approaches for low-light enhancement. 
    (a) HDR-based methods leverage raw or linear HDR images to preserve the physical radiance of the scene while avoiding the nonlinear distortions introduced by standard ISP pipelines. Neural rendering is performed directly on the raw domain, followed by image signal processing (ISP) to produce the final rendered image. 
    (b) Decomposition-based methods explicitly model low-light degradations (e.g., noise, blur) and tone mapping as separate components. These factors are jointly considered during neural rendering to improve the recovery of structure and luminance under challenging illumination conditions.
}
    \label{fig:low-light}
\end{figure}

Drantal-NeRF is a framework designed to achieve high-fidelity 3D reconstruction by removing aliasing artifacts from degraded NeRF renderings. In the first stage, a NeRF model is trained from multi-view degraded inputs while fine-tuning a diffusion model \cite{b147} to restore anti-aliased HQ images. In the second stage, a Controllable Feature Wrapping module adjusts VAE decoder \cite{b22,b148} features based on encoder outputs to improve fidelity. A patch-based discriminator is used for adversarial training, aligning the anti-aliased outputs with the real GT distribution and improving multi-view consistency and reconstruction fidelity.

\subsection{Enhancement in 3D LLV}
3D Enhancement refers to techniques that improve the visual quality of multi-view 3D scene representations, focusing on preserving fine structural details, refining surface textures, and enhancing reconstruction under low-light conditions. Unlike 2D enhancement, which is limited to pixel-level improvements in single images, 3D enhancement must simultaneously ensure geometric consistency across views and maintain spatial coherence. This necessitates designs that jointly correct both structural and photometric aspects of the scene representation, as categorized in the taxonomy shown in Fig. \hyperref[fig:tax]{6}.

\subsubsection{Low-light Enhancement}
Low-light enhancement in this context requires spatially and temporally coherent modeling of lighting, reflectance, and visibility. Recent studies address these challenges through HDR-based, Retinex-based, and degradation modeling-based approaches \cite{b115, b116, b117, b123}.

As shown in Fig. \hyperref[fig:low-light]{11}, HDR-based methods operate directly on raw or linear HDR inputs to preserve the physical radiance of the scene and avoid nonlinear distortions introduced by standard ISP pipelines. On the other hand, both Retinex-based and degradation modeling-based methods follow a decomposition-based strategy that explicitly separates low-light degradations (noise, blur) from tone-mapping. This design enables the network to recover structural and photometric fidelity more effectively during neural rendering under challenging illumination conditions \cite{b122}.

\paragraph{HDR-based}
This group of methods leverages raw or linear HDR images to preserve the physical radiance of the scene, avoiding the nonlinear distortions introduced by conventional ISP pipelines. RawNeRF \cite{b115} supervises NeRF training directly on linear raw data, enabling HDR novel view synthesis with user-defined exposure and tone mapping. 
Bright-NeRF \cite{b123} models the spectral response of individual color channels under low-light conditions and applies view-dependent color correction matrices to reduce illumination-induced distortions. While highly effective at preserving luminance and color fidelity, these methods require access to raw sensor data and knowledge of camera-specific parameters.

\paragraph{Retinex-based}
Retinex-based methods are grounded in the theory that an observed image can be decomposed into intrinsic reflectance and external illumination components. LLNeRF \cite{b116} factorizes the radiance field into a view-independent reflectance and a view-dependent lighting term, enabling consistent reconstruction under varying illumination. 

Heterogeneous LLNeRF \cite{b117} extends this formulation to unsupervised training using inputs with heterogeneous exposure conditions. 

Thermal-NeRF \cite{b121} incorporates thermal imagery as an auxiliary modality to improve illumination estimation, particularly in extremely dark environments. These approaches achieve robust generalization to lighting variations but rely on explicit decomposition modules, which require careful design.

\paragraph{Degradation modeling-based}
This class of methods focuses on modeling and disentangling compound degradation factors common in low-light imagery, including visibility loss, sensor noise, and camera motion blur. 

Aleth-NeRF \cite{b120} introduces a Concealing Field that attenuates visibility according to a global darkness parameter, enabling robust rendering in varying low-light conditions. These methods avoid explicit radiometric priors and directly handle real-world degradations, making them well-suited for in-the-wild applications.

Ambient-NeRF \cite{b118} models low-light degradation as a Hadamard product of the normal-light scene and a learned ambient illumination tensor, inspired by the Phong reflection model \cite{b206}. Without requiring prior darkness parameters, it jointly learns the radiance and ambient components via a lightweight MLP and employs a custom loss function. Accelerated by hash encoding \cite{b111} and FullyFusedMLP \cite{b207}, it enables fast and physically interpretable low-light NeRF reconstruction.

Gaussian in the Dark \cite{b119} extends the 3DGS framework by introducing a learned camera response function and light-aware Gaussian primitives, allowing the model to infer exposure conditions and lighting levels. 

LuSh-NeRF \cite{b122} assumes an implicit order of degradation and proposes a sequential correction pipeline, combining a Scene-Noise Decomposition module with a Camera Trajectory Prediction module.

\subsubsection{Detail Enhancement}
\begin{figure}[t]
    \centering
    \includegraphics[width=0.9\linewidth]{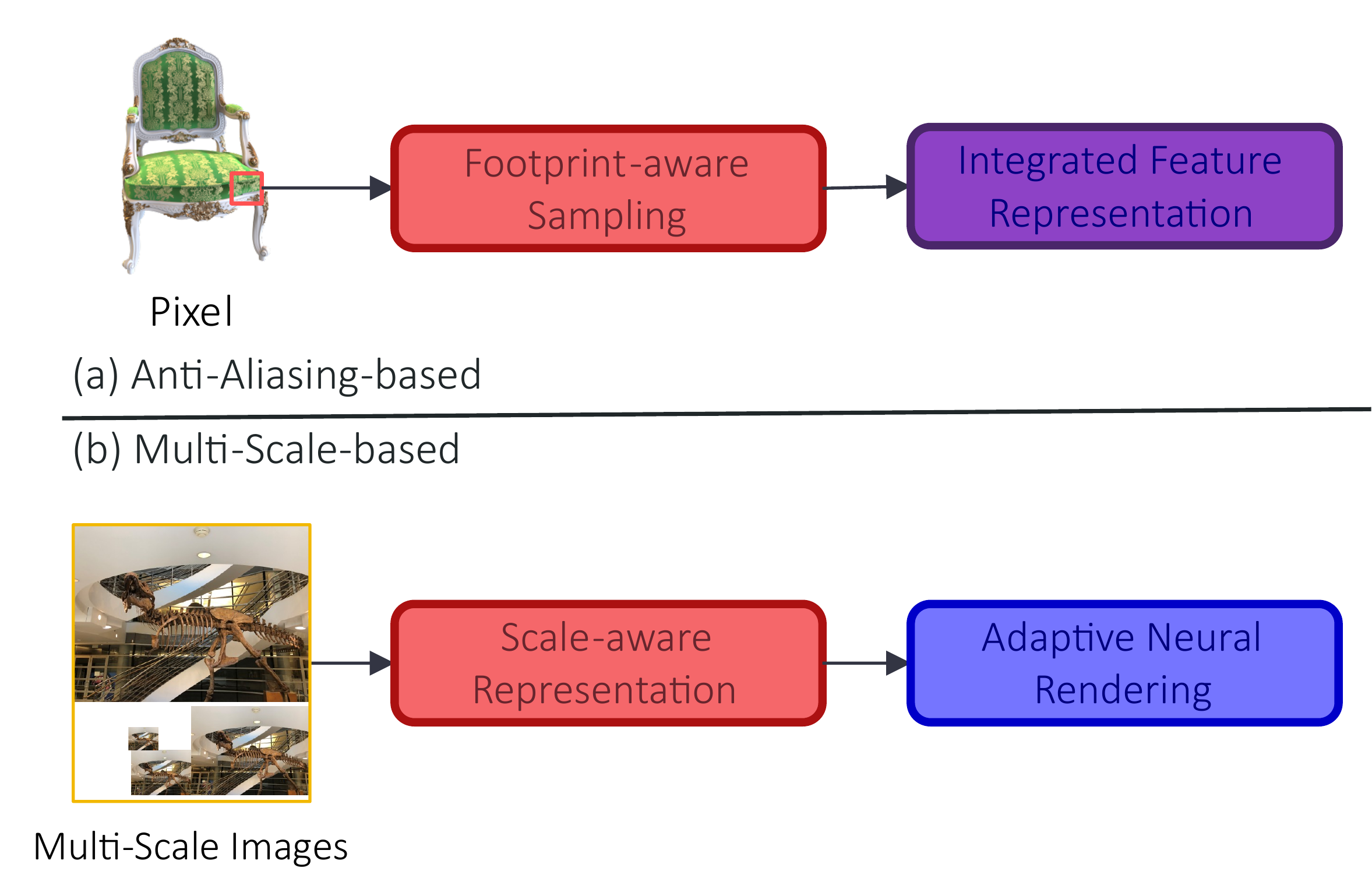}
    \caption{Comparison of two representative detail enhancement strategies in neural rendering. (a) Anti-aliasing-based method from mip-NeRF~\cite{b40}, which leverages conical frustums and computes Integrated Positional Encoding (IPE) to produce smooth and stable reconstructions. (b) Multi-scale-based method that constructs scale-aware features from multi-resolution inputs and enhances details via adaptive rendering.}
    \label{fig:detail}
\end{figure}
Detail enhancement is crucial for photorealistic scene reconstruction in neural rendering, where multi-scale observations from varying viewpoint distances or resolutions often cause blurring in close-up views and aliasing in distant views, leading to structural inconsistency and visual instability \cite{b138}.

To address this, recent research introduces multi-scale architectures and scale-aware sampling techniques that either suppress aliasing by modeling the pixel footprint and sampling geometry or dynamically adjust network capacity based on spatial scale variations. As shown in Fig. \hyperref[fig:detail]{12}, these approaches are typically categorized into (a) anti-aliasing-based methods, which compute expected positional encodings over conical frustums to suppress artifacts at thin structures or high-frequency regions, and (b) multi-scale-based methods, which progressively refine features by integrating information across multiple spatial resolutions.

\paragraph{Anti-Aliasing-based}
Mip-NeRF \cite{b53} replaces NeRF’s point sampling with cone tracing, modeling each ray as a conical frustum. This region is approximated as a multivariate Gaussian, and integrated positional encoding (IPE) is introduced to represent both the position and the spatial extent of each sample. This allows the network to handle scale variance in footprint size and suppress aliasing in distant regions.

Mip-NeRF 360 \cite{b50} extends this framework to unbounded outdoor scenes. It explicitly separates foreground and background modeling and introduces inverse depth space sampling to increase sampling density in distant regions. This leads to more stable learning in large-scale environments. Geometric distortion is mitigated through regularization.

Zip-NeRF \cite{b133} systematically analyzes the imbalance of sampling along the depth axis and introduces the notion of Z-aliasing, which arises from overly dense samples in distant regions. To address this, Zip-NeRF applies a view-dependent level-of-detail control strategy that adjusts sampling frequency and network capacity based on camera distance, enabling consistent anti-aliasing across varying depth ranges.

Tri-MipRF \cite{b208} introduces triangle-based mipmaps for efficient aliasing removal in NeRF. It projects ray cones \cite{b53} onto triangular regions and constructs a multi-scale representation that more accurately approximates anisotropic pixel footprints. An efficient cone-casting strategy compatible with hash encoding \cite{b111} is integrated to enable area-aware sampling. By leveraging triangle-based kernels for adaptive level selection and radiance filtering, Tri-MipRF achieves twice the training speed of mip-NeRF while preserving high-frequency details and improving rendering stability.

Mip-Splatting \cite{b134} applies mipmapping principles to 3DGS. Each Gaussian splat is precomputed at multiple levels of detail, and the appropriate level is selected during rasterization based on the pixel footprint. This reduces splat overlap or omission and improves rendering stability and visual quality.

\paragraph{Multi-Scale-based}
Bungee-NeRF \cite{b135} introduces a progressive training strategy that begins with a shallow base block modeling distant views. As training proceeds, new blocks are incrementally added to capture fine details at closer views. High-frequency components in the positional encoding are gradually activated, allowing stable learning of fine-scale geometry without early overfitting.

PyNeRF \cite{b139} trains multiple model heads at different spatial resolutions and selects the appropriate head based on pixel footprint. This allows for efficient inference while reducing aliasing artifacts.

Mip-VoG \cite{b136} proposes a hierarchical image-pyramid-based feature structure. Feature weights are dynamically adjusted based on sampling location and scene scale, enabling real-time rendering with consistent detail and photometric coherence across multiple resolutions.

MS-GS \cite{b137} dynamically adjusts splat size and frequency based on spatial scale and depth. Distant objects are rendered using broad, low-frequency splats, while nearby objects are modeled with compact, high-frequency splats. This achieves effective level-of-detail control while minimizing aliasing artifacts and maintaining visual fidelity.

HoGS \cite{b138} addresses the limitation of Cartesian coordinates in 3DGS by introducing homogeneous coordinates. Based on projective geometry, this representation enables scale-invariant modeling and accurate rendering of both near and distant objects in large, unbounded outdoor environments. HoGS improves geometric stability and consistency across varying depths and perspectives.

\subsubsection{Texture Enhancement}
\begin{figure}[t]
    \centering
    \includegraphics[width=\linewidth]{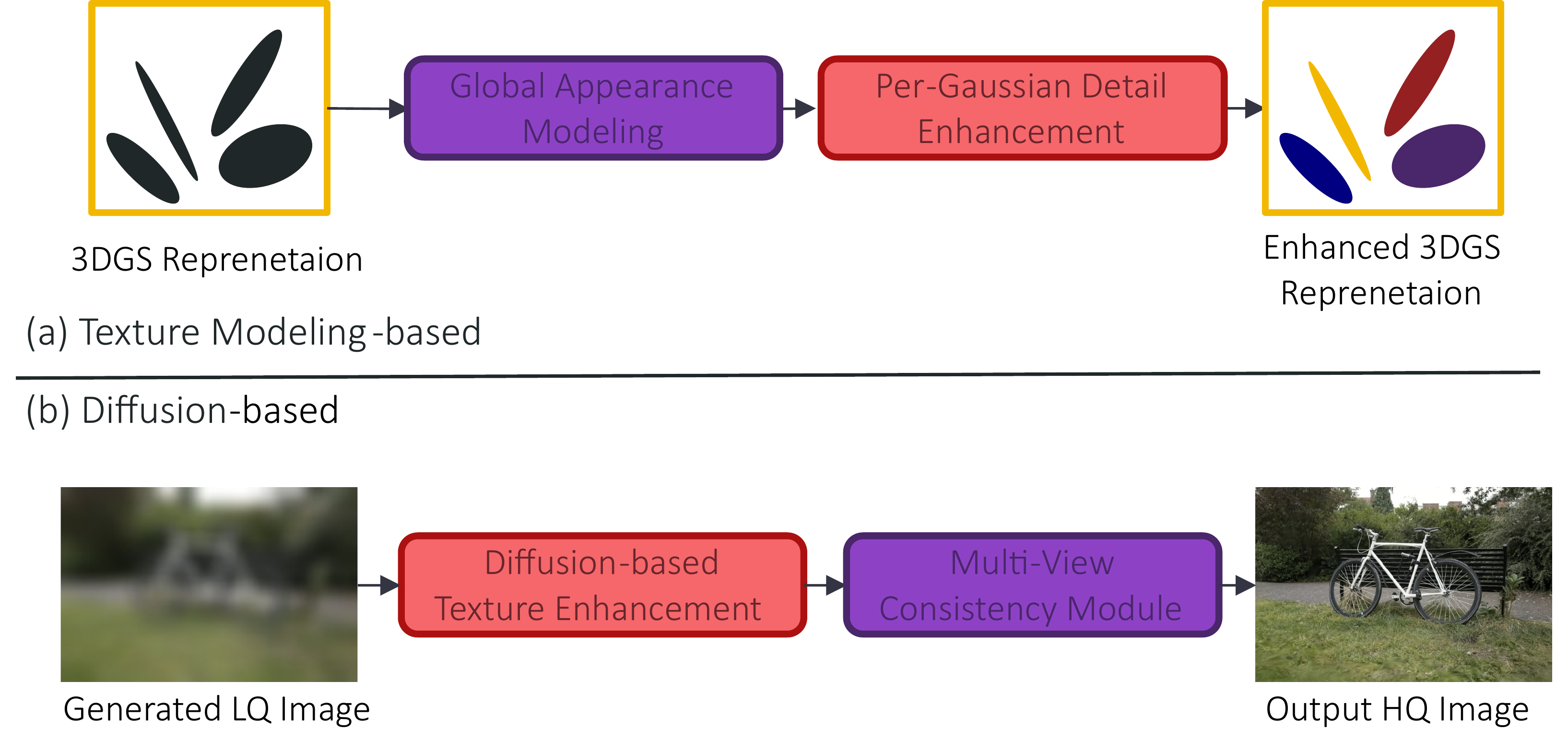}
    \caption{Illustrates two representative approaches for texture enhancement in 3D neural rendering.
    (a) Texture modeling-based: Enhances the 3DGS representation by refining primitive-level appearance parameters to produce perceptually richer textures.
    (b) Diffusion-based: Applies a diffusion model \cite{b112} to refine low-quality textures, followed by a consistency module that enforces multi-view alignment in the rendered output.
}
    \label{fig:texture}
\end{figure}
Texture enhancement plays a crucial role in neural rendering for realistic scene reconstruction, particularly in restoring high-frequency details from sparse viewpoints or LR images \cite{b129, b130}. Traditional approaches often struggle to reproduce realistic textures due to limited expressivity and inconsistencies across views \cite{b129}. Recent studies address these challenges by proposing a variety of methodologies, which can be broadly classified into diffusion-based post-enhancement and conditioned texture modeling approaches, as illustrated in Fig. \hyperref[fig:texture]{13}.

\paragraph{Diffusion-based}
These methods enhance texture quality via diffusion models applied to rendered images or latent spaces, without altering the underlying geometry.

LATTE3D \cite{b128} leverages a 3D-aware diffusion prior \cite{b112} to perform localized texture refinement over coarse geometry. By combining user-painted guidance with 3D geometric embeddings in a transformer-based architecture, it achieves prompt-consistent and detail-preserving texture generation in the text-to-3D pipeline.

3DGS-Enhancer \cite{b127} refines pretrained 3DGS scenes by rendering multi-view images and applying a 2D latent diffusion model \cite{b112} to enhance texture appearance. The updated outputs are back-projected to fine-tune appearance attributes of the Gaussians, improving texture fidelity while maintaining 3D geometry.

3DENHANCER \cite{b129} proposes epipolar-aware cross-view attention to guide multi-view latent diffusion \cite{b112}, explicitly modeling geometric correspondences across views. This design enhances multi-view texture consistency and enables robust synthesis under sparse and misaligned inputs.

\paragraph{Texture modeling-based}
These methods directly integrate texture reasoning into the 3D representation by conditioning color, opacity, or learned appearance fields on spatial or semantic priors.

Texture-GS \cite{b209} introduces per-Gaussian texture descriptors to explicitly disentangle structural and textural components within 3D Gaussian Splatting. These descriptors are optimized through a hierarchical pipeline and refined via view-consistent alignment modules, enabling high-frequency detail preservation and consistent texture rendering across views.

Textured-GS \cite{b130} enhances standard 3DGS by introducing spatially defined color and opacity using SH. This modification allows each Gaussian to represent intra-splat variation in appearance, significantly improving realism without increasing parameter count.

Textured Gaussians \cite{b132} replaces traditional per-Gaussian color parameters with learnable RGBA appearance maps. These appearance maps are optimized via multi-view rendering losses and enable spatially dense, view-consistent texture representations across scenes.

%% file: sections/Experimental_Setup.tex
\section{Experimental Setup} \label{Dataset}

\subsection{Dataset}
This section provides a comprehensive overview of widely adopted datasets that serve as standard benchmarks in recent literature on 3D reconstruction, neural rendering, and novel view synthesis. These datasets are essential for consistent evaluation and comparison across diverse experimental settings in 3D LLV. Table \hyperref[tab:nerf_datasets]{1} summarizes their key characteristics, including scene type, number of images, resolution, and available supervision.

\definecolor{RealType}{RGB}{213, 236, 234}
\definecolor{SyntheticType}{RGB}{255, 230, 230}

\newcommand{\linkicon}[1]{\href{#1}{\faExternalLink}}

\newcommand{\typeboxR}{\raisebox{-.2ex}{%
\tcbox[colback=RealType, colframe=white, boxsep=0pt,
left=2pt, right=2pt, top=0.2pt, bottom=0.2pt,
arc=1pt, boxrule=0pt]{\textbf{R}}}}
\newcommand{\typeboxS}{\raisebox{-.2ex}{%
\tcbox[colback=SyntheticType, colframe=white, boxsep=0pt,
left=2pt, right=2pt, top=0.2pt, bottom=0.2pt,
arc=1pt, boxrule=0pt]{\textbf{S}}}}

\newcommand{\typeboxRS}{\typeboxR\!\typeboxS}
\definecolor{MotionStatic}{RGB}{230, 230, 230}
\definecolor{MotionDynamic}{RGB}{200, 220, 255}

\newcommand{\motionboxS}{\raisebox{-.2ex}{%
\tcbox[baseline, colback=MotionStatic, colframe=white, boxsep=0pt,
  left=2pt, right=2pt, top=0.5pt, bottom=0.5pt, arc=1pt, boxrule=0pt]{\textbf{S}}}}

\newcommand{\motionboxD}{\raisebox{-.2ex}{%
\tcbox[baseline, colback=MotionDynamic, colframe=white, boxsep=0pt,
  left=2pt, right=2pt, top=0.5pt, bottom=0.5pt, arc=1pt, boxrule=0pt]{\textbf{D}}}}

\begin{table*}[t]
\centering
\small
\caption{
Summary of widely-used datasets for 3D reconstruction and novel view synthesis,
including dataset name, year, scene type (\typeboxR: real, \typeboxS: synthetic), number of scenes,
number of images per scene, resolution, motion type (\motionboxS: static, \motionboxD: dynamic),
with clickable links on the right.
}
\label{tab:nerf_datasets}
\begin{tabular*}{0.9\textwidth}{l c c c c c c >{\centering\arraybackslash}p{2cm}}

\toprule
\textbf{Dataset} & \textbf{Year} & \textbf{Type} & \textbf{\#Scenes} & \textbf{\#Images} & \textbf{Resolution} & \textbf{Motion} & \textbf{Link} \\
\midrule

DTU \cite{b151} & 2014 & \typeboxR & 124 & 49 or 64 & $1600 \times 1200$ &\motionboxS & \linkicon{https://roboimagedata.compute.dtu.dk/?page_id=36} \\

Tanks and Temples \cite{b48} & 2017 & \typeboxR & 21 & 100--400 & $1920 \times 1080$ & \motionboxS & \linkicon{https://www.tanksandtemples.org/} \\

Deep Blending \cite{b49} & 2018 & \typeboxR & 19 & 12--418 & $1288 \times 816$ & \motionboxS & \linkicon{https://github.com/Phog/DeepBlending} \\

LLFF \cite{b51} & 2019 & \typeboxR & 8 & 20--62 & $6032 \times 3024$ & \motionboxS & \linkicon{https://github.com/Fyusion/LLFF} \\

Stereo Blur Dataset \cite{b152} & 2019 & \typeboxR & 135 & Video & $1280 \times 720$ & \motionboxD & \linkicon{https://shangchenzhou.com/projects/stereoblur/} \\

NeRF-synthetic \cite{b1} & 2020 & \typeboxS & 8 & 400 & $800 \times 800$ & \motionboxS & \linkicon{https://github.com/bmild/nerf} \\

BlendedMVS \cite{b52} & 2020 & \typeboxR\typeboxS & 113 & 150--200 & $2048 \times 1536$ & \motionboxS & \linkicon{https://github.com/YoYo000/BlendedMVS} \\

NSVF Synthetic \cite{b54} & 2020 & \typeboxS & 8 & 400 & $800 \times 800$ &   \motionboxS & \linkicon{https://github.com/facebookresearch/NSVF} \\

HyperNeRF \cite{b153} & 2021 & \typeboxR & 7 & Video & $1920 \times 1080$ & \motionboxD & \linkicon{https://hypernerf.github.io/} \\

Deblur-NeRF \cite{b65} & 2022 & \typeboxRS & 31 & 27--53  &  $600 \times 400$ & \motionboxD & \linkicon{https://github.com/limacv/Deblur-NeRF} \\

NeRF in the Dark \cite{b115} & 2022 & \typeboxRS & 5 & 25--200 & $6000 \times 4000$ & \motionboxS & \linkicon{https://bmild.github.io/rawnerf} \\

Mip-NeRF 360 \cite{b50} & 2022 & \typeboxR & 9 & 100--330 &  $4096 \times 3286$ & \motionboxS & \linkicon{https://jonbarron.info/mipnerf360/} \\

RTMV \cite{b55} & 2022 & \typeboxR & 2000 & 150 & $1600 \times 1600$ & \motionboxS & \linkicon{https://github.com/antony-wu/rtmv} \\

iPhone Dataset \cite{b155} & 2022 & \typeboxR & 14 & Video & $720 \times 960$ & \motionboxD & \linkicon{https://github.com/apple/ml-hdrnet} \\

Objaverse \cite{b154} & 2023 & \typeboxS & 800K+ & 3D Object  & - & \motionboxS & \linkicon{https://objaverse.allenai.org/} \\

\bottomrule
\end{tabular*}
\end{table*}

The \textbf{DTU dataset} \cite{b151} consists of multi-view still images captured under controlled indoor lighting conditions. It contains 124 scenes, each composed of 49 to 343 HR images. For each scene, GT 3D point clouds are provided via structured-light scanning. Camera intrinsics and poses are included, enabling precise geometric alignment. 

The \textbf{Tanks and Temples dataset} \cite{b48} comprises real-world scenes captured in diverse indoor and outdoor environments for 3D reconstruction. Each scene contains approximately 100 to 400 images, with GT 3D meshes obtained via laser scanning. The dataset features structural complexity, varying illumination, and large-scale spatial diversity, making it suitable for challenging benchmarks. It serves as a standard benchmark for multiview stereo (MVS) and neural rendering methods. 

\textbf{Deep Blending} is a dataset \cite{b49} for few-shot novel view synthesis based on sparse real-world inputs. Each scene consists of 10 to 418 images with camera poses estimated via COLMAP and coarse 3D proxy geometry. Due to sparse viewpoint coverage, it is used to evaluate performance under limited observation conditions. Although dense GT meshes are not provided, point clouds are available for visual quality assessment. The dataset supports research on sparse-view enhancement \cite{b130, b132}.

The \textbf{LLFF dataset} \cite{b51} includes forward-facing real-world scenes captured along semi-circular camera paths. Each scene contains approximately 20 to 62 images, with camera poses recovered via COLMAP. All cameras face toward the center of the scene, ensuring consistent viewing direction. It includes complex textures, non-uniform lighting, and fine-scale geometry, providing a suitable testbed for evaluating generalization performance.

The \textbf{Stereo Blur Dataset} \cite{b152} contains stereo image pairs in which one view is blurred while the other remains sharp. It consists of 135 video sequences and a total of 20,637 stereo frames with motion and defocus blurs. The dataset is split into 98 training sequences with 17,319 image pairs and 37 testing sequences with 3,318 pairs. GT disparity maps are provided, enabling depth-guided deblurring and alignment.

The \textbf{NeRF-synthetic dataset} \cite{b1} contains eight synthetic scenes rendered using Blender \cite{b180} with complex geometry and materials. Each scene consists of 100 to 300 images distributed across 360-degree or hemispherical viewpoints. A standard 3-way split is used with 100 images each for training, validation, and testing. Camera intrinsics and poses are provided, and all images are noise-free with consistent illumination. This dataset is widely adopted for benchmarking neural rendering.

\textbf{BlendedMVS dataset} \cite{b52} consists of 113 scenes that combine real backgrounds with rendered 3D objects for domain-diverse MVS learning. Each scene contains 150 to 200 HR images with COLMAP-reconstructed camera poses and depth maps. The dataset is split into 106 training scenes and 7 validation scenes. It reflects real-world visual statistics while maintaining full 3D supervision.

The \textbf{NSVF Synthetic dataset} \cite{b54} includes eight synthetic scenes with complex geometry, fine structures, and challenging lighting. Each scene contains approximately 400 HR images, rendered with global illumination effects. It is commonly split by index, with views 0 to 99 used for training, 100 to 199 for validation, and 200 to 399 for testing. Each image includes a depth map and camera pose. 

The \textbf{HyperNeRF dataset} \cite{b153} consists of dynamic scenes with non-rigid deformations captured using monocular or stereo video. Each sequence contains several hundred frames, along with COLMAP-reconstructed camera poses and LiDAR-based or inferred depth maps. It includes fine-grained temporal shape variations and realistic motion patterns. HyperNeRF is commonly used to evaluate dynamic scene reconstruction performance in real-world environments.

\textbf{Deblur-NeRF dataset} \cite{b65} introduces a dataset for learning to reconstruct sharp 3D scenes from blurred inputs. It consists of 31 dynamic scenes, each containing 27 to 53 multi-view images synthesized using Blender. Motion blur is simulated by interpolating perturbed camera poses, while defocus blur is rendered via built-in depth-of-field functionalities. The dataset supports both camera motion and defocus blur analysis, enabling evaluation of deblurring in neural rendering frameworks.

\textbf{NeRF in the Dark dataset} \cite{b115} provides 5 real-world scenes captured in low-light conditions using RAW images. Each scene includes 25 to 200 images with resolution of 
$ 6000 \times 4000$ , covering a wide dynamic range. It enables the study of neural rendering from photon-limited, noise-corrupted observations. The dataset includes corresponding exposure metadata and GT tone-mapped references, and is widely used to benchmark low-light view synthesis methods.

\textbf{Mip-NeRF 360 dataset} \cite{b50} contains nine real-world scenes captured along 360-degree camera trajectories in both indoor and outdoor settings. Each scene includes several hundred HR images, covering a wide range of scales from near objects to distant backgrounds. COLMAP-based camera intrinsics and poses are provided. The dataset emphasizes unbounded scene modeling, floating-point precision, and background consistency. It is commonly used to evaluate rendering stability, spatial generalization, and large-scale reconstruction.

The \textbf{RTMV dataset} \cite{b55} consists of approximately 2,000 synthetic 3D scenes rendered via path tracing to produce over 300,000 HR images. Each scene includes 150 to 160 multiview images with GT depth maps and camera metadata. Scenes vary in shape, material, and lighting, with diverse spatial layouts. Although no standard split is defined, subsets are often used for large-scale supervised training. RTMV is a benchmark for scalable view synthesis, SR, and general-purpose neural field learning.

The \textbf{iPhone dataset} \cite{b155} contains monocular videos of 14 dynamic real-world scenes captured using handheld smartphones. Each video includes several hundred frames, with synchronized views from secondary devices used for evaluation. COLMAP-based camera poses and LiDAR-based depth are provided. Seven additional sequences are reserved for multiview validation. The dataset reflects real-world motion, occlusion, and illumination variations, making it suitable for testing dynamic neural rendering under realistic conditions.

\textbf{Objaverse dataset} \cite{b154} provides over 800,000 textured 3D objects along with associated metadata such as mesh structure, texture maps, and semantic tags. Each object can be rendered into dozens of viewpoints using standard rendering pipelines. A subset of 100,000 objects and 1.2 million rendered images is typically used for training and validation. The dataset supports object-centric neural rendering and text-to-3D modeling. Its massive scale and variety make it suitable for training generative and reconstruction-based 3D models.

\subsection{Metrics}
We categorize and summarize representative evaluation metrics used to assess the visual quality and consistency of generated 3D content. These metrics are essential for quantitatively validating the performance of 3D LLV methods under various degradation conditions. The representative metrics are categorized by reference type and scoring direction, as shown in Table~\ref{tab:3dllv_metrics}.

Among the most commonly used full-reference metrics are PSNR \cite{b56}, SSIM \cite{b57}, and LPIPS \cite{b58}. PSNR measures the pixel-wise reconstruction accuracy by computing the logarithmic ratio between the maximum possible pixel value and the mean squared error. SSIM evaluates structural similarity by comparing luminance, contrast, and structure between the rendered 2D images and GT images. LPIPS, on the other hand, measures perceptual similarity using deep features extracted from pretrained neural networks \cite{b181,b182}, capturing high-level semantic differences that may not be reflected in traditional pixel-based metrics.

In addition to full-reference methods, a wide range of no-reference metrics have been proposed to evaluate visual quality without requiring GT data. These include traditional statistical approaches such as NIQE \cite{b157}, PIQE \cite{b158}, BRISQUE \cite{b159}, and LIQE \cite{b160}, as well as learning-based methods like RankIQA \cite{b161}, MetaIQA \cite{b162}, MUSIQ \cite{b163}, CLIP-IQA \cite{b164}, and MANIQA \cite{b165}. These metrics are particularly useful for assessing generalization in real-world scenarios where GT data is often unavailable. They provide insight into perceptual quality and distortion levels in uncontrolled or degraded conditions, making them suitable for evaluating 3D rendering outputs under adverse inputs.

To assess temporal consistency across video frames or multi-view sequences, the temporal Optical Flow (tOF) metric \cite{b156} is employed. tOF quantifies motion coherence by measuring discrepancies in optical flow fields between consecutive frames. High temporal stability indicates that the generated outputs maintain consistent motion patterns and structural alignment over time, which is critical in dynamic scenes and real-time applications.
Together, these metrics offer a comprehensive framework for evaluating both spatial quality and temporal stability of 2D rendered images from 3D LLV representations. Their complementary nature ensures robust performance assessment across a wide range of rendering conditions and application scenarios.

\begin{table}[!t]
\centering
\caption{Summary of evaluation metrics used in 3D LLV.\\
Arrows (\faArrowUp/\faArrowDown) indicate whether higher or lower values correspond to better visual quality.}
\label{tab:3dllv_metrics}
\begin{tabularx}{\linewidth}{l c c c}
    \toprule
    \textbf{Category} & \textbf{Metric and Citation} & \textbf{Scoring Direction} \\
    \midrule

    \multirow{3}{*}{\makecell{Full-reference}}
    & PSNR~\cite{b56} & \faArrowUp \\
    & SSIM~\cite{b57} & \faArrowUp \\
    & LPIPS~\cite{b58} & \faArrowDown \\

    \midrule

    \multirow{9}{*}{\makecell{\centering No-reference}}
    & NIQE~\cite{b157} & \faArrowDown \\
    & PIQE~\cite{b158} & \faArrowDown \\
    & BRISQUE~\cite{b159} & \faArrowDown \\
    & LIQE~\cite{b160} & \faArrowDown \\
    & RankIQA~\cite{b161} & \faArrowUp \\
    & MetaIQA~\cite{b162} & \faArrowUp \\
    & MUSIQ~\cite{b163} & \faArrowUp \\
    & CLIP-IQA~\cite{b164} & \faArrowUp \\
    & MANIQA~\cite{b165} & \faArrowUp \\

    \midrule

    \multirow{1}{*}{\makecell{\centering Temporal}}
    & tOF~\cite{b156} & \faArrowDown \\

    \bottomrule
\end{tabularx}
\end{table}

\subsection{Training strategy}
\subsubsection{SR}
In 3D neural rendering, SR training typically begins by constructing a LR 3D scene representation based on NeRF or 3DGS, using multi-view LR images as input. The 2D LR inputs are commonly generated by applying bicubic downsampling to 2D HR images with scale factors such as $\times2$, $\times4$, or $\times8$, depending on the experimental setting \cite{b5,b6,b28,b19}. The corresponding 2D HR images are provided directly from the original dataset and serve as GT supervision.

In image-based strategies, pretrained SISR models \cite{b10,b11,b20, b33, b112} are applied to each rendered LR image. The resulting SR outputs are then fed back into the 3D optimization process to enhance high-frequency details and texture fidelity.

In video-based strategies, a sequence of LR views rendered from the 3D model is enhanced using a pretrained VSR model \cite{b30, b27}. The resulting HR video sequence is used to update the 3D representation with improved temporal and view-consistent detail.

\subsubsection{Deblurring}
In neural rendering methods for motion and defocus deblurring, training is performed without directly using GT sharp images.Instead, the model learns by comparing the rendered blurry output with the observed blurry input, as illustrated in Fig.~\ref{fig:degradation-aware-rendering}(a). For motion blur, sharp views are rendered along the exposure trajectory and temporally aggregated to synthesize blur, which is then aligned with the input image during optimization, as shown in Fig. \hyperref[fig:deblur]{8 (a), (b)}. Training data includes both real motion blur captured during camera exposure and synthetic blur generated by averaging temporally aligned sharp frames \cite{b70}. Event-based approaches \cite{b77,b78,b79,b80} incorporate high-temporal-resolution event streams to provide structural cues that guide learning. Trajectory-based methods  \cite{b68, b69, b71, b72} explicitly model camera motion and jointly optimize scene geometry and motion parameters. For defocus blur, neural rendering models simulate blur by expanding spatial contributions according to the depth-dependent CoC, which is derived from the scene depth, aperture size, and focus distance \cite{b83,b89}. The model minimizes the discrepancy between the rendered defocused output and the observed input image, as illustrated Fig. \hyperref[fig:deblur]{8 (c)}. Some methods further estimate per-view focus and aperture settings and incorporate CoC-based masks to enhance structure preservation in in-focus regions.

\subsubsection{Weather Degradation Removal}
The datasets are composed of multi-view images or videos degraded by weather conditions such as haze, rain, or water droplets, collected either from real-world captures or through synthetic generation. While GT clean images may be available, they are not directly used for supervision but serve as references for evaluation. Most methods adopt neural rendering frameworks that integrate weather-specific degradation models to jointly learn scene geometry and degradation components, as shown in Fig.\hyperref[fig:degradation-modeling]{3 (b)}. Training is conducted in an unsupervised or self-supervised manner, primarily guided by reconstruction loss between degraded input views and rendered outputs. High-level regularization techniques are employed to maintain physical consistency and structural coherence \cite{b34,b41,b43,b46}. Some approaches further enhance learning stability by masking out localized degradations such as occlusions or adherent droplets \cite{b62,b63}. Overall, these methods enable the separation of weather-induced degradation and the recovery of clear 3D scenes without relying on explicit GT supervision, as shown in Fig. \hyperref[fig:degradation-aware-rendering]{2 (a)}.

\subsubsection{Restoration}
Restoration methods for neural rendering aim to reconstruct high-fidelity 3D scenes from degraded multi-view inputs affected by noise, blur, compression artifacts, or LR. Datasets are typically constructed by applying synthetic degradations to high-quality multi-view data or by collecting real-world captures that naturally exhibit such degradations\cite{b99,b98}. Although clean GT images may be available, they are generally used only for evaluation purposes or as auxiliary signals, rather than for direct supervision. Models are trained in a supervised or weakly supervised manner, guided by restoration priors derived from diffusion models \cite{b108,b109,b110,b147}, generative adversarial networks, or degradation-aware modules. Learning strategies frequently involve decoupling geometry from appearance, addressing cross-view inconsistencies, and refining radiance fields through residual correction mechanisms or adversarial objectives. Many approaches model degradation as a learnable distribution rather than relying on explicit formulations, thereby providing robustness across diverse types of corruption. The overarching objective is to enhance the resilience of 3D reconstruction pipelines to real-world imperfections while maintaining structural and visual fidelity.

\subsubsection{Enhancement}
Low-light enhancement methods train neural rendering models using multi-view images captured under low illumination with severe noise. The datasets are constructed either by capturing real low-light scenes with short exposure times or by synthetically darkening normal-light images and adding noise. While normal-light images may exist, they are not used as direct supervision signals. Instead, the models learn to decompose the input into illumination, reflectance, and noise components, enabling reconstruction of scenes under normal lighting conditions \cite{b115, b116, b117, b123}, as shown in Fig.~\ref{fig:degradation-modeling}(b). Illumination effects are modeled through an additional representational space, where various mechanisms, such as Retinex-based intrinsic decomposition and tone mapping operators, are employed to adjust or compensate for lighting variations across views. Training is predominantly unsupervised, with the primary objective being the faithful reconstruction of geometrically consistent scenes under normal illumination across multiple views.

Detail enhancement methods aim to recover fine structures by addressing blur and aliasing artifacts caused by multi-scale observations at varying distances. Datasets are typically constructed from real or synthetic scenes captured at multiple resolutions and viewpoints, encompassing varying levels of detail from near to far. Although GT images may exist, most methods adopt unsupervised training using photometric consistency and multi-view alignment without direct supervision. During training, techniques such as multi-scale representation \cite{b135,b139}, frustum-based sampling \cite{b53}, and frequency-aware filtering \cite{b133} are applied to preserve spatial details \cite{b53, b50, b133}. Models often allocate more representational capacity to nearby regions or progressively refine scene representations according to viewpoint or resolution. These strategies enable consistent reconstruction of scene details across scales within the same environment.

Texture enhancement methods aim to restore high-frequency surface details while maintaining consistency across multi-view inputs. Datasets typically consist of real or synthetic scenes with multi-view RGB images and camera poses \cite{b127, b130, b132}. Some approaches initialize with coarse 3D geometry or use sparse viewpoints, and may further incorporate text prompts or external priors. Although GT textures may be available, they are generally not used for direct supervision. Instead, training is conducted under unsupervised or weakly supervised paradigms guided by generative priors \cite{b112}. Learning strategies are designed to mitigate texture misalignment through view-consistent diffusion, texture mapping, or latent feature synchronization. To preserve both spatial detail and cross-view structural coherence, models incorporate techniques such as texture sharing, mask-based refinement, and multi-view attention.

%% file: sections/Future_Directions.tex
\section{Future Directions} \label{Future}
Blind LLV has recently garnered significant attention in the 2D domain due to its ability to restore images under various degradation conditions without requiring explicit information about the degradation kernels. Kernel estimation-based blind SR methods \cite{b166, b167} estimate the degradation kernel from LR inputs and utilize it to reconstruct high-frequency details, enabling accurate SR. Self-enhancement-based blind deblurring approaches \cite{b168} leverage repetitive patterns and structural priors within the image to effectively remove blur. Diffusion-based restoration methods \cite{b169} employ powerful generative models to recover images under diverse degradation conditions.
These approaches are designed under the assumption that degradation kernels or conditions are unknown, which aligns well with the challenges encountered in 3D LLV tasks requiring multi-view restoration. For instance, in 3D scenes reconstructed from sparse point clouds generated by SfM, or in scenarios with degradation inconsistencies across views, blind restoration strategies can serve as effective solutions. Building on their success in 2D tasks, these methods demonstrate strong potential for extension and adaptation to 3D LLV problems.

3D LLV for dynamic scenes has shown significant potential through recent advances in dynamic scene modeling. Models such as 4D Gaussian Splatting \cite{b183}, Grid4D \cite{b184}, SC-GS \cite{b185}, D-NeRF \cite{b186}, and NeRF-DS \cite{b187}, although originally designed for rendering and reconstruction, offer core architectural components that are well-suited for LLV integration. These methods incorporate mechanisms such as time-conditioned scene representation \cite{b186, b187}, spatial-temporal consistency \cite{b183, b184}, non-rigid motion modeling \cite{b186}, specular appearance handling \cite{b187}, and sparse control for editable reconstruction \cite{b185}, all of which are valuable for restoration in degraded dynamic environments.
Currently, Dynamic Scene LLV has been primarily applied to motion deblurring \cite{b76, b77, b78, b79, b80}, focusing on mitigating temporal degradations such as motion blur and frame misalignment.
Building on these foundations, the architectures of dynamic scene models provide a structural basis for extending LLV tasks to other time-aware applications such as SR, restoration, and enhancement. These directions offer distinct advantages for addressing temporally varying degradations that static LLV methods cannot effectively resolve, positioning Dynamic Scene LLV as a promising framework for robust 3D restoration in real-world dynamic environments.

Sparse-view 3D LLV directly aligns with the goals of LLV, as both deal with degraded or insufficient input information. Recent works have proposed effective solutions to handle sparse observations, such as dense point cloud generation \cite{b188}, depth-aligned modeling \cite{b189}, and multi-View feature integration \cite{b190}.
These methods not only enhance the geometric fidelity of 3D reconstructions but also provide transferable components for LLV tasks. For instance, depth alignment can support denoising and deblurring, while semantic features and view-aware embeddings are beneficial for super-resolution and localized enhancement. Therefore, the core techniques used in sparse-view reconstruction offer a promising foundation for improving LLV performance under limited input conditions.

Real-Time LLV has become an essential direction for deploying low-level vision models in latency-sensitive environments such as mobile platforms, AR/VR systems, and real-time robotics. In the 2D domain, recent methods have explored model compression through knowledge distillation and pruning \cite{b194}, architectural optimization using reparameterization techniques \cite{b195}, and task-specific branching designs for efficient inference \cite{b196}. These approaches aim to reduce computational cost while maintaining high restoration performance. In the 3D domain, real-time LLV remains in its early stages, but initial efforts include lightweight point cloud denoising using fixed trigonometric encodings \cite{b197} and grid-based ground completion leveraging 2D–3D hybrid architectures for LiDAR \cite{b198}. Such strategies lay the groundwork for fast and responsive 3D restoration, highlighting the potential of real-time LLV systems in practical 3D scenarios.

All-in-One Restoration and Enhancement aims to process various types of degradation such as noise, blur, compression artifacts, and low-light conditions within a single unified framework. In recent 2D research, multiple approaches have been proposed, including diffusion-based restoration models \cite{b199}, degradation-aware state-space models \cite{b200}, prompt-conditioned convolutional architectures \cite{b192}, gated latent-space representations for multi-degradation handling \cite{b201}, and conditional diffusion models that automatically detect and adapt to degradation types \cite{b191}. In addition to these, residual transformer-based frameworks utilizing grid structures \cite{b202} and prompt–ingredient-oriented architectures for unified inference \cite{b203} have also demonstrated strong potential for generalized restoration. These methods are designed to operate without explicit knowledge of degradation conditions, achieving generalization by sharing model parameters, injecting conditional priors, and modulating latent representations.
Such strategies are well aligned with the challenges of multi-view 3D LLV, where degradations are often heterogeneous across views and ground-truth degradation information is unavailable. By enabling a single model to jointly handle diverse degradations across different perspectives, these approaches can enhance both spatial consistency and restoration efficiency in 3D reconstruction. Therefore, all-in-one architectures are expected to play a pivotal role in the future of 3D LLV by providing a scalable and robust framework for real-world multi-view restoration under complex and uncertain degradation scenarios. 

Interactive Restoration or Enhancement introduces a human-in-the-loop paradigm, where user intent is actively incorporated into the restoration process. Some methods enable users to select specific spatial regions through mouse interaction and provide natural language instructions to control localized restoration effects \cite{b177, b204}. Others inject prompt embeddings into intermediate layers of the model to modulate the restoration strength, type of degradation to address, or stylistic preferences based on textual cues \cite{b192, b191}. Semantic scene descriptions can further guide the restoration network by conditioning it on high-level contextual understanding \cite{b205}, enabling alignment with user intent even in the absence of low-level visual cues. These mechanisms offer fine-grained control and adaptability. When extended to 3D LLV, they have strong potential to support view-specific correction, spatiotemporal prioritization, and interactive resolution of multi-view inconsistencies, suggesting a promising future direction for developing user-adaptive 3D restoration systems.

%% file: sections/Conclusion.tex
\section{Conclusion} \label{Conclusion}
This paper reviewed the current progress in 3D LLV, which aims to restore and reconstruct high-quality 3D scenes from degraded 2D inputs. We focused on how LLV tasks such as SR, deblurring, Weather Degradation Removal, Enhancement, and Restoration are applied within 3D rendering frameworks like NeRF and 3DGS.

We explained the challenges posed by real-world degradations, including LR, noise, and blur, and how they affect the quality and consistency of 3D reconstruction. To support future development, we introduced common datasets and evaluation methods used in this field.

3D LLV plays an important role in applications such as autonomous driving, AR/VR, and robotics, where accurate and reliable 3D information is needed even under poor imaging conditions. By integrating restoration processes directly into the rendering pipeline, 3D LLV makes it possible to recover detailed and consistent 3D scenes.

This survey summarized recent research trends, identified remaining challenges, and highlighted the importance of developing more robust 3D LLV frameworks that can operate reliably under various degradation scenarios. We hope this paper provides a useful foundation for further research in achieving high-quality 3D perception in real-world environments.